\documentclass[final,3p,times,twocolumn]{elsarticle}

\usepackage{amsmath,amssymb,amsfonts}
\usepackage{comment}
\usepackage{float}
\usepackage{enumitem} 

\usepackage{hyperref}

\usepackage[dvipsnames]{xcolor}

\journal{Neural Networks}

\newcommand{\brav}[1]{{\color{black} #1}}
\usepackage{soul}

\newcommand{\mas}[1]{{\color{teal} #1}}

\def\sname{\textit{EmbBERT}}

\begin{document}

\begin{frontmatter}

\title{EmbBERT: Attention Under 2 MB Memory}

\author[a]{Riccardo Bravin} 
\author[a]{Massimo Pavan} 
\author[a]{Hazem Hesham Yousef Shalby} 
\author[a]{Fabrizio Pittorino}
\ead{fabrizio.pittorino@polimi.it}
\author[a]{Manuel Roveri} 
\affiliation[a]{organization={Department of Electronics, Information and Bioengineering, Politecnico di Milano},
            addressline={Via Ponzio
34/5}, 
            city={Milano},
            postcode={20133}, 
            country={Italy}}

\begin{abstract}

Transformer architectures based on the attention mechanism have revolutionized natural language processing (NLP), driving major breakthroughs across virtually every NLP task. However, their substantial memory and computational requirements still hinder deployment on ultra-constrained devices such as wearables and Internet-of-Things (IoT) units, where available memory is limited to just a few megabytes. To address this challenge, we introduce \textbf{\sname{}}, a tiny language model (\textbf{TLM}) architecturally designed for extreme efficiency. The model integrates a compact embedding layer, streamlined feed-forward blocks, and an efficient attention mechanism that together enable optimal performance under strict memory budgets. Through this redesign for the extreme edge, we demonstrate that highly simplified transformer architectures remain remarkably effective under tight resource constraints. \sname{} requires only \textbf{2~MB} of total memory, and achieves accuracy performance comparable to the ones of  state-of-the-art (SotA) models that require a $\mathbf{10\times}$ memory budget. Extensive experiments on the curated \textbf{TinyNLP} benchmark and the \textbf{GLUE} suite confirm that \sname{} achieves competitive accuracy, comparable to that of larger SotA models, and consistently outperforms downsized versions of BERT and MAMBA of similar size. Furthermore, we demonstrate the model’s resilience to 8-bit quantization, which further reduces memory usage to just \textbf{781~kB}
, and the scalability of the \sname{} architecture across the sub-megabyte to tens-of-megabytes range. Finally, we perform an ablation study demonstrating the positive contributions of all components and the pre-training procedure. All code, scripts, and checkpoints are publicly released to ensure reproducibility: \url{https://github.com/RiccardoBravin/tiny-LLM}.

\end{abstract}

\end{frontmatter}


\section{Introduction}
\label{sec:introduction}

The proliferation of Internet-of-Things (IoT) systems and the availability of off-the-shelf, energy-efficient pervasive technologies have sparked a growing demand for intelligent on-device computation~\citep{iot}. Tiny devices like microcontrollers and wearables now play a central role in applications ranging from smart homes to industrial automation~\citep{wearables}. Despite significant advances in their memory and processing capabilities, these tiny devices still operate under stringent constraints, making very challenging the design and development of machine (ML) and deep learning (DL) models meant to be executed on these devices~\citep{MCUNet,Eff_NN_for_embedded}. This need pushed the research field of Tiny Machine Learning (TinyML), which enables efficient ML/DL inference on devices characterized by limited memory (e.g., less than 2~MB memory), low processing power (e.g., less than 100~MHz CPUs), and strict energy budgets~\citep{tinyML}, hence allowing tiny devices to locally process data, reducing latency, improving the real-time responsiveness of smart applications, and enhancing privacy by keeping sensitive data on-device.

While TinyML has made strides in areas like keyword spotting~\citep{on_device_kw_spotting,customizable_kw_spotting}, image classification~\citep{mobilenet_v2,micro_net_img_rec}, and object detection~\citep{EfficientDet_obj_det}, deploying \emph{natural language processing (NLP)} models on tiny devices remains a significant challenge. Modern Large Language Models (LLMs) based on attention mechanisms, such as BERT~\citep{BERT}, XLNet~\citep{XLNet}, DistilBERT~\citep{DistilBERT}, SpanBERT~\citep{SpanBERT}, ALBERT~\citep{ALBERT}, RoBERTa~\citep{RoBERTa} and State-Space-Models (SSMs) such as MAMBA~\citep{MAMBA}, rely on millions or even billions of parameters and substantial memory resources to achieve SotA accuracy across a wide range of NLP tasks. Even scaled-down variants like MobileBERT (25.3M parameters)~\citep{MobileBERT} are orders of magnitude too large for deployment on microcontrollers with less than 2~MB memory budgets.

To fill this gap, we introduce \emph{\sname{}}, a \emph{Tiny Language Model (TLM)}, specifically designed to operate on tiny devices, and under the stringent 2~MB memory budget. 
In particular, \emph{\sname{}} comprises a novel TLM architecture optimized for microcontroller units and other resource-constrained devices. 

Redesigning the architectural components of transformer-based architectures for working under strict memory budgets, \sname{}, with a memory requirement of only 2~MB, achieves SotA performance comparable to the ones of models with $10\times$ its memory requirements, such as BERT-Tiny~\citep{tinyBERT}. Even when compared with other models specifically adapted to work within the 2~MB constraints, \sname{} resulted in being the model with the best performance. 

Our contributions include:
\begin{enumerate}
    \item \emph{\sname{}}: A new TLM specifically designed for tiny devices, combining efficiency and effectiveness.
    \item \emph{Memory and Computational Analysis}: An analytical evaluation of the memory usage and computational complexity of \sname{} and its components, providing a useful tool to evaluate weights and activations memory trade-offs required to operate within tiny device constraints.
    \item \emph{Analyses on the quantization of the model and scalability of the architecture:} An analytical evaluation of the effects of a post-training 8-bit quantization scheme on the accuracies and memory requirements of the model, and an analysis on the scalability of the \sname{} architecture across the sub-megabyte to tens-of-megabytes range. 
    \item \emph{Custom Benchmark}: A specialized benchmark tailored to assess the NLP capabilities of TLMs, enabling consistent and fair evaluations in resource-constrained environments.
\end{enumerate}

The remainder of this paper is organized as follows. Section~\ref{sec:relatedwork} reviews recent work on model compression and training techniques tailored for resource-constrained platforms, setting the stage for our contributions. Section~\ref{sec:EmbBERT} provides a detailed description of \sname{}, our proposed model for NLP in tiny devices, and includes precise calculations of memory requirements for its layers. In Section~\ref{sec:setup}, we outline the experimental setup, including the training procedures and datasets used to validate our approach. Section~\ref{sec:results} presents a comprehensive evaluation of our model on the TinyNLP benchmark suite and GLUE datasets, comparing downscaled versions of BERT and MAMBA, and highlighting the significant performance improvements achieved by \sname{}. In Section~\ref{sec:quant_n_scale}, we provide an analysis of the effects that quantization and scaling have on the model accuracies and memory requirements, while Section~\ref{sec:ablation} delves into an ablation study to assess the individual contributions of the architectural modules within \sname{} and the pre-training procedure to its overall performance. Finally, Section~\ref{sec:conclusions} concludes the paper by discussing the experimental findings, exploring future research directions, and considering the broader implications of deploying TLMs in tiny devices.


\section{Related Work}
\label{sec:relatedwork}
 
This section introduces the SotA for Small Language Models and the main techniques applied to reduce their memory and computational demands.

\paragraph{Small Language Models}\hfill \break
Several models, including BERT-Tiny~\citep{tinyBERT}, NanoBERT~\citep{NanoBERT}, MobileBERT~\citep{MobileBERT}, ConvBERT~\citep{ConvBERT}, and I-BERT~\citep{i-BERT} are considered at the SotA for Small Language Models, each leveraging a different compression strategy such as distillation, architectural simplification, or quantization. 

\textit{BERT-Tiny}~\citep{tinyBERT} employs a shallow architecture with fewer layers and reduced hidden dimensions with respect to conventional BERT architectures, trading accuracy for a smaller size and lower computational costs (4.4~M parameters corresponding to a 20~MB memory footprint when also considering activations). It employs a knowledge distillation approach from a larger teacher model to a compact student model, achieving competitive accuracy through task-specific and data-augmented distillation. 

\textit{NanoBERT}~\citep{NanoBERT} represents, in current literature, the smallest model built for tiny devices. It manages to achieve it results by introducing a \emph{Nano Embedder}, a clever mechanism that aims to reduce the memory footprint of the standard embedder. Still, with weights in the range 700-800~k, experimental results confined to a few datasets and missing code and checkpoints, it falls behind other more prominent publications. \textit{MobileBERT}~\citep{MobileBERT}, on the other hand, employs bottleneck structures and a teacher-student training to compress BERT, retaining high accuracy at the expense of a higher number of weights~(25.3~M), making it better suited for edge devices than actual tiny devices.

\textit{ConvBERT}~\citep{ConvBERT}, with it small size of 14~M parameters, introduces convolutional operations in self-attention to reduce computational complexity and efficently capture local dependencies. \textit{I-BERT}~\citep{i-BERT} employs integer-only quantization, guaranteeing that all computations, including softmax and matrix multiplications, rely on integer arithmetic. This allows for significant improvements on memory and energy efficiency without compromising accuracy, despite its larger size of 355~M parameters.

\paragraph{Advanced Training and Pre-Training Techniques}\hfill \break
Training schemes beyond traditional masked language modeling (MLM)~\citep{BERT} play a crucial role in enhancing the efficiency and performance of smaller models. Techniques like \textit{ELECTRA} generative-discriminative training~\citep{ELECTRA} and \textit{LoRA} low-rank updates~\citep{LoRA} enable fine-tuning with reduced trainable parameters, optimizing computational demands while retaining accuracy. Additionally, data augmentation and synthetic text generation~\citep{data_augmentation, synthetic_data} expand training datasets, enhancing knowledge transfer and improving small-model generalization.

These methods are particularly relevant for small language models, as they provide mechanisms to further reduce resource requirements while maintaining competitive performance. Nevertheless, the largest advantages of these techniques are obtained during training, a phase that is usually not carried out on the devices.


\paragraph{Quantization and Knowledge Distillation}\hfill \break
Quantization, a cornerstone of model compression, converts weights and activations to lower-precision formats, such as 8-bit integers, drastically reducing memory usage and improving inference efficiency~\citep{quantization}. Integer arithmetic, as demonstrated by I-BERT~\citep{i-BERT}, aligns well with embedded hardware capabilities. Knowledge distillation~\citep{knowledge_distillation}, employed in models like Bert-Tiny and MobileBERT, further reduces model size by training smaller networks to replicate larger teacher models' behavior.

\paragraph{Exploring Alternative Architectures} \hfill \break
Beyond transformers, alternative architectures like recurrent neural networks (RNNs) with state-space models (SSMs) represent a viable option for tiny devices. For example, \textit{MAMBA}~\citep{MAMBA}, with 140~M parameters, leverages RNNs to mitigate the attention quadratic complexity, offering efficient text generation solutions that could be even tailored for environments with limited parallelism capabilities due to it inherent recursive structure.

\vspace{5mm}

Despite substantial progress in model compression, 
most existing approaches still fail to meet the stringent memory and computational constraints of ultra-low-resource devices. 
In contrast, our work introduces a complete memory-aware redesign that jointly minimizes parameter counts and activation usage, rather than merely scaling down existing models. By combining compact embeddings, simplified attention, and feed-forward structures, \sname{} achieves remarkable results with just a 2~MB footprint - enabling deployment on devices far smaller than any supported by prior models.


\begin{figure*}[t]
    \centering
    \includegraphics[width=0.75\linewidth]{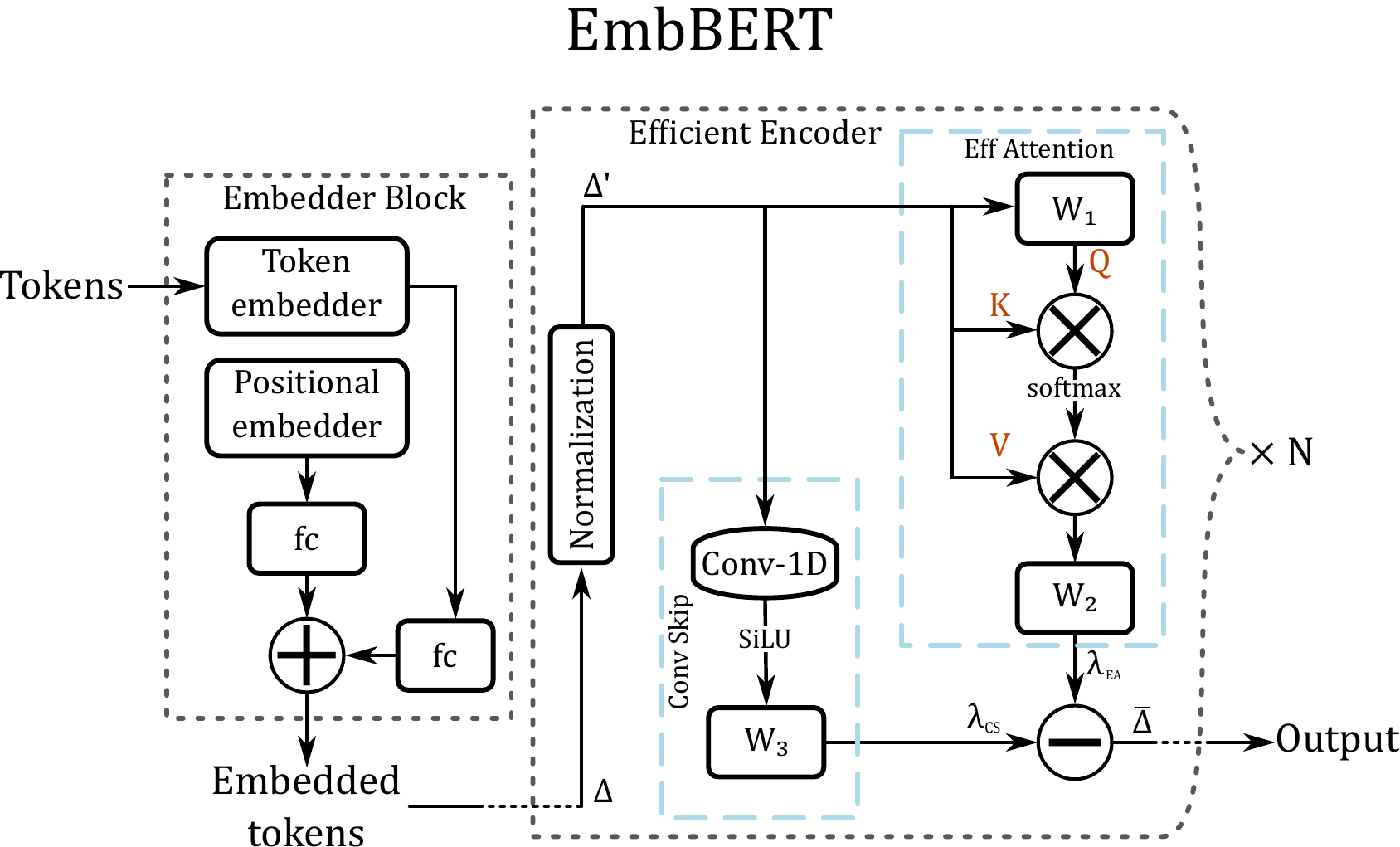}
    \caption{Overview of the \sname{} architecture, a specialized Language Model designed for tiny devices and extreme resource constraints. The architecture features an Embedder block and multiple Efficient Encoder with an Efficient Attention block and a Convolutional Skip Connection block. Instead of a final sum to aggregate the two Efficient Encoder paths, \sname{} uses a weighted difference with learnable weights for enhanced performance.}
    \label{fig:EmbBert_overview}
\end{figure*}

\section{\sname{}: A Novel Language Model for Tiny Devices}
\label{sec:EmbBERT}

\sname{} is a compact and efficient language model designed specifically for deployment in memory-constrained environments. Unlike conventional approaches that prioritize performance at the expense of footprint, \sname{} strikes a fine balance by leveraging targeted architectural optimizations. These innovations make it particularly suitable for real-world natural language processing tasks in TinyML scenarios, where resource efficiency is paramount. The \sname{} architecture is represented in Fig.~\ref{fig:EmbBert_overview}.

\subsection{The \sname{} Architecture}

\sname{} achieves its efficiency through a redesign of traditional encoder-based BERT models, prioritizing lightweight attention mechanisms and memory-aware component selection. This architecture focuses on minimizing memory usage without sacrificing accuracy, surpassing other lightweight models within similar constraints. 


\sname{} comprises two main modules (delineated by dashed gray lines): a \textit{Nano Embedder Block}, responsible for generating compact yet expressive token representations called $\delta$s; and a sequence of $N$ \textit{Efficient Encoder blocks} integrating efficient attention, convolutional layers, and weighted difference aggregation to process embeddings with minimal memory overhead.
Additionally, an optional \textit{Output Block} can be appended to adapt the architecture for specific downstream regression, classification or generative tasks.
In the following sections, we explore the functionality and design of these core modules, highlighting the architectural innovations that make \sname{} a SotA solution for tiny devices.

\subsubsection{The Embedder block}

The Embedder block in \sname{} is adapted from the Nano Embedder introduced in ~\citep{NanoBERT} (whose memory requirements are analyzed in~\ref{appendix:memory}). 

The Embedder block generates a $d$-dimensional representation~$\delta$ for each token in a vocabulary of size $v$ using its token embedder. These embeddings are positioned in a $d$-dimensional space, reflecting semantic similarity among tokens. To capture word order, the model includes a positional encoding component that maps each token position (up to a maximum window size $\ell$, usually referred to as sentence length) into a~$d$-dimensional vector, ensuring that token order is understood alongside their meaning. Unlike the original Nano Embedder~\citep{NanoBERT}, our Embedder block employs a learned positional embedding, built on the same principles as the token embedder. This approach not only enhances representation quality but also reduces memory requirements compared to using a fixed positional embedder. Additionally, segment embeddings, similar to those used in BERT, assign distinct $d$-dimensional vectors to differentiate between input segments (e.g., sentences), enabling the model to interpret relationships across boundaries. 

A key part of the Embedder block memory efficiency is in its use of dense layers after the embeddings. Specifically, the Embedder maps input tokens into a reduced space of dimension $r_d$, and the fully connected layer projects them into the desired $d$-dimensional space. This results in a $d \times \ell$ matrix $\Delta$ of embedded tokens $\delta$s for a sentence length~$\ell$.
This approach significantly reduces the Embedder size while maintaining sufficient representational power for classification tasks. By decreasing memory requirements for the embedding layer, the Embedder Block frees up space for other parts of the network, making it a crucial enabler for running language models on devices with limited memory.

\brav{
\subsubsection{The Efficient Encoder}

\begin{table*}[t]
    \caption{Formulas for calculating the weights and activation sizes of the blocks and components of \sname{}.}
    \begin{center}
        \begin{tabular}{|c | c c|}
        \hline
        \textbf{Layers} & \textbf{Weights} & \textbf{Activations} \\
        \hline \hline
        \textbf{Embedder block}           & \(r_d \cdot (v + \ell + 2 d ) + 2 d\)
                                        & \(r_d \cdot \ell + 2 d \cdot \ell\) \\ \hline \hline
        \textbf{Norm layer}          & \(2 d\)
                                        & \(2 d \cdot \ell\) \\
        \textbf{Eff. Attention block}    & \(2 d^2 \) 
                                        & \(2 d \cdot \ell + \ell^2 \)\\
        \textbf{Conv Skip block}    & \(d^2 \cdot \alpha + k \cdot d \cdot \alpha\)
                                        & \(d \cdot \ell (2 + \alpha)\)\\
                                        \hline
        \textbf{Efficient Encoder} & \(2d + 2 d^2 + d^2 \cdot \alpha + k \cdot d \alpha\)
                                        & \(\max ( 2 d \cdot \ell + \ell^2 ; \  d \cdot \ell (2 + \alpha )) \)\\
        \hline
        \end{tabular}
    \end{center}
    \label{table:w&a_noMAMBA}
\end{table*}

The core of \sname{} is a sequence of $N$ \textit{Efficient Encoder} blocks (right gray dashed block of Fig.~\ref{fig:EmbBert_overview}). Each block processes the embedded token matrix $\Delta$ through a normalization layer followed by two parallel paths: an Efficient Attention mechanism  and a Convolutional Skip Connection. 
The Efficient Attention mechanism developed in \cite{efficient_attention}, is a lightweight alternative to Standard Attention, which can provide similar accuracy results while at the same time reducing by roughly 50~\% the required amount of parameters and the computational demands with respect to standard attention \cite{efficient_attention}. More detailed comparisons on the complexities of these two attention mechanism can be found in \ref{appendix:memory}.

Our choice of efficient attention is motivated by the fact that, in the sub-2MB end-to-end memory regime (weights + activations) targeted here, the dominant bottleneck is 
the fixed per-layer footprint of projections and intermediate tensors. Removing the key/value projections reduces both parameters and activations, freeing memory that can be reallocated to higher-impact capacity (e.g., wider embeddings or additional mixing) under the same global budget. By contrast, many efficient-attention variants are designed to reduce the $O(L^2)$ dependence on sequence length and often require extra projections/state that are not competitive under our strict memory accounting~\cite{effattn}. For this reason, we adopt a softmax-based, minimal-state attention block and leave broader comparisons to long-context efficient-attention mechanisms to future work.

Let $\Delta' = \text{Norm}(\Delta)$ be the normalized input. The Efficient Attention pathway computes:
\[
Q = W_1(\Delta'), \qquad K = \Delta', \qquad V = \Delta',
\]
and produces
\[
\Delta_{EA} = W_2\!\left(
\text{Softmax}\!\left(
\frac{Q K}{\sqrt{d}}
\right) V
\right).
\]
We adopt a single attention head ($h=1$), as higher head counts increase activation memory and have been shown to be less effective at this scale~\citep{exploring_transformer_sizes}.

In parallel, the Convolutional Skip Connection applies a depthwise 1D Conv layer with kernel size $k$ and channel expansion factor $\alpha$, acting as a lightweight local token mixer:
\[
\Delta_{CS} =
W_3\cdot\!\bigl(
\text{SiLU}(\text{Conv1D}(\Delta'))
\bigr),
\]
where padding and reshaping preserve the sequence length $L$.

The block output is obtained by a learned weighted combination of the two pathways:
\[
\overline{\Delta}
=
\lambda_{EA}\,\Delta_{EA}
-
\lambda_{CS}\,\Delta_{CS},
\]
and is passed either to the next Efficient Encoder or to the final output module.

By integrating single-head attention, depthwise convolutions, and a lightweight aggregation step, the Efficient Encoder substantially reduces both parameter count and activation footprint. This enables flexible trade-offs among embedding size, token mixing capacity, and memory usage, making \sname{} deployable within strict 1--2\,MB TinyML memory budgets.}

\subsection{Computing memory requirements of \sname{}}

An accurate computation of the total memory requirements for running \sname{} models is required in severely resource-constrained settings. The total memory, $M_{tot}$, is the sum of the memory needed for its weights and activations, and can be expressed as:
\begin{equation*}
    M_{tot} = (W_{emb} + N \cdot W_{enc}) \cdot P_w + \max(A_{emb},\,A_{enc}) \cdot P_a,
    \label{eq:embedder_params}
\end{equation*}
where $W_{emb}$ and $W_{enc}$ are the weights of the embedder and encoder, respectively, and $A_{emb}$ and $A_{enc}$ are their respective activations. $P_w$ and $P_a$ denote the precision (in bits) used to store weights and activations. 

The formulas used to compute the memory required for the weights and activations of the components of \sname{} are reported in Table~\ref{table:w&a_noMAMBA}. The next subsections are dedicated to an in-depth analysis of their memory and computational requirements, while we reported the way in which these requirements are derived in \ref{appendix:memory}.

\subsubsection{The Embedder block}

The Embedder block  of \sname{} is composed of 5 smaller components: token embedder of size \( v \times r_d \), positional embedder of size \( \ell \times r_d \), segment embedder of size \( 2 d \) and two fully connected layers of size $r_d \times d$ (where $v$ is the vocabulary size; $\ell$ the embedding size, and $r_d$ the reduced embedding size). Together, they result in a total parameter count:
\begin{equation*}
W_{emb} = r_d \cdot (v + \ell + 2 d) + 2 d,
\end{equation*}
achieving a reduction in parameter size of almost a $d / r_d$ factor with respect to a standard Embedder. However, the total activation size $A_{emb}$ required by the Embedder block slightly increases due to the added projection step, resulting in: 
\begin{equation*}
    A_{emb} = r_d \cdot \ell + 2 d \cdot \ell.
\end{equation*}

During inference, the Embedder block performs as a first step $\ell + 2\ell \cdot r_d$ memory accesses for token and position encoding, followed by $4\ell \cdot r_d \cdot d$ memory accesses, $2\ell \cdot r_d \cdot d + \ell \cdot d$ summations, and $2\ell \cdot r_d \cdot d $ multiplications for the linear layers, and in case segment embeddings are needed another \(\ell + 2d + \ell \cdot d \) accesses and \(\ell \cdot d\) summations.

\subsubsection{The Efficient Encoder}
We now proceed to analyze the memory usage and computational complexity of the three primary blocks that define the Efficient Encoder of \sname{}.

The normalization layer’s weights consist just of two vectors of size $d$ representing the averages and standard deviations required for normalization. This block also has a minimal activation size, requiring only $2d \cdot \ell$ values. The operations within this layer necessitate \(2d \cdot (\ell + 1)\) memory accesses, along with $d \cdot \ell$ summations and multiplications. 

From a memory perspective, the Efficient Attention block requires $2d^2$ weights and has an activation size of $2\ell \cdot d + \ell^2$. 
While its memory demands are relatively small, this efficiency comes at the cost of a high computational complexity. Due to the various matrix multiplications and the softmax operation, the block requires: \(4\ell \cdot d^2 + 4\ell^2 \cdot d + 2\ell^2\)  memory accesses, \(2\ell \cdot d^2 + 2\ell^2 \cdot d + \ell^2 + \ell \cdot d \) summations and  \(2\ell \cdot d^2 + 2\ell^2 \cdot d + 2\ell^2\) multiplications. 

\brav{
Compared to standard scaled dot-product attention, Efficient Attention eliminates the explicit key and value projection matrices. While a conventional attention layer requires $3hd^2$ projection parameters, Efficient Attention reduces this to $hd^2$, yielding:
\[
\frac{hd^2}{3hd^2} = \tfrac{1}{3},
\]
i.e., a $66\%$ reduction in projection parameters. In addition, removing the key and value projections eliminates two matrix multiplications per head, reducing both memory accesses and arithmetic operations during inference~\citep{efficient_attention}.
}

The Convolutional Skip Connection block, compared to the feed-forward block that typically follows the standard attention layer and that is not present in \sname{}, features reduced memory requirements for the weights. This reduction arises from replacing the fully connected expansion layer with a Conv-1D expansion layer. Consequently, this block requires:
\(d^2 \cdot \alpha + k \cdot d \cdot \alpha\) weights and \(d \cdot \ell (2 + \alpha)\) values for activations. Its computational requirements are \(d \cdot \ell (k+4 + 2 d \cdot  \alpha) \) memory accesses, \(d \cdot \ell (k + 2 + d \cdot \alpha)\) sums and \(d \cdot \ell (k + 7 + d \cdot \alpha)\) multiplications.

For the aggregation step, the block requires only \(2 d \cdot \ell\) 
multiplications and memory accesses, with half as many summations. This step does not introduce any additional weights or activations.

Summing up, the Efficient Encoder of \sname{} requires:
\begin{equation*}
    W_{enc} = 2 d^2 + d^2 \cdot \alpha + k \cdot d \cdot \alpha  + 2d
\end{equation*}
weights and has a total activation size of:
\begin{equation*}
    A_{enc} = \max ( 2 d \cdot \ell + \ell^2 ; \  d \cdot \ell (2 + \alpha )).
\end{equation*}
Overall, the self-attention mechanism structured weighting of sequence elements enables robust contextual representations, ensuring that tokens are influenced by all relevant inputs regardless of distance. 

All \sname{} layers processing is performed using the full floating precision at 4 bytes, but they can be easily quantized for improved performance and memory occupation as discussed in section \ref{subs:quantization}.

\subsection{Selecting the Architectural Parameters for \sname{}}

The selection of optimal architectural parameters for the \sname{} model is a critical step in ensuring both performance and memory efficiency. Key parameters such as vocabulary size~$v$, sentence length~$\ell$, and embedding dimension~$d$ play a significant role in determining the model’s memory usage and overall effectiveness. These parameters heavily influence the embedder memory occupation and cannot be significantly reduced without adversely affecting model accuracy.

To adapt the model to the 2MB memory constraint, the first step involves identifying the smallest feasible values for~$v$ and~$\ell$ that maintain acceptable performance. This study evaluated~$v$ within the range of 2048 to 16384 to maintain the model expressiveness, while~$\ell$ was chosen between 256 and 1024. The choice of~$\ell$ was informed by the average sentence lengths in the selected datasets and their interplay with vocabulary size.

The embedding dimension~$d$ and the reduced embedding dimension~$r_d$ were then tuned. While these dimensions can be scaled down more aggressively, reducing~$d$ below 128 results in significant performance degradation, as shown in~\citep{exploring_transformer_sizes} and section~\ref{subsec:scaling}. For~$r_d$, values between 16 and 32 were found to yield optimal results, aligning with the findings of~\citep{NanoBERT}.

Finally, other structural parameters such as the scaling factor~$\alpha$, the convolutional kernel size ~$k$ and the number of layers~$N$ were fine-tuned to balance memory constraints with performance objectives, ensuring the model operated effectively within the given limitations.

\begin{table*}[htbp]
    \caption{\textbf{Model hyperparameters}. Columns represent: vocabulary size $v$, sentence length $\ell$, embedding dimension $d$, reduced embedding dimension $r_d$, forward expansion $\alpha$, number of attention heads $h$, SSM state size $d_s$, convolution kernel size $k$, and number of layers.}
    \begin{center}
    
        \begin{tabular}{|c|c c c c c c c c c|}
            \hline
            \textbf{Model} & $v$ & $\ell$ & $d$ & $r_d$ & $\alpha$ & $d_s$ & $k$ & $h$ & $N$  \\
            \hline \hline
            Embedder                   & 8192  & 256 & 320 & 32 & /    & / & /  & / & 1 \\
            Embedder~+~conv                 & 8192  & 256 & 320 & 32 & /    & / & 16 & / & 1 \\
            BERT(2MB)~\cite{BERT}      & 2048  & 256 & 80  & /  & 2    & / & /  & 2 & 2 \\
            MAMBA(2MB)~\cite{MAMBA}    & 2048  & 256 & 64  & /  & 1    & 6 & 4  & / & 5 \\
            \brav {NanoBERT(2MB)~\cite{NanoBERT}}    & 8192 & 256 & 90 & 16 & 2 & / & / & 2 &	2 \\
            \textbf{\sname{} (ours)}           & 8192  & 256 & 128 & 16 & 1    & / & 32 & 1 & 4 \\
            \hline \hline
            \textcolor{gray}{\emph{BERT-Tiny} (20MB)~\cite{tinyBERT}}   & \textcolor{gray}{32768} & \textcolor{gray}{512} & \textcolor{gray}{128} & \textcolor{gray}{/}  & \textcolor{gray}{2}    & \textcolor{gray}{/} & \textcolor{gray}{/}  & \textcolor{gray}{2} & \textcolor{gray}{2}\\
            \hline
        \end{tabular}

    \end{center}
    \label{table:models_params_and_memory}
\end{table*}

\begin{table*}[htbp]
    \caption{\textbf{Model resource usage}. Columns represent weights, activations, total size, number of additions (sums), multiplications (mults), and memory accesses.}
    \begin{center}
        \begin{tabular}{|c|c c c | c c c|}
            \hline
            \textbf{Model} & \textbf{Weights} & \textbf{Activations} & \textbf{Total size} & \textbf{Sums} & \textbf{Mults} & \textbf{Mem. accesses} \\
            \hline \hline
            Embedder                   & 293 K     & 164 K     & 1,826 MB & 11 M  & 5 M   & 11 M \\
            Embedder~+~conv            & 298 K     & 164 K     & 1,848 MB & 13 M  & 7 M   & 12 M \\
            BERT(2MB)~\cite{BERT}      & 289 K     & 213 K     & 2,008 MB & 137 M & 69 M  & 69 M \\
            MAMBA(2MB)~\cite{MAMBA}    & 220 K     & 265 K     & 1,941 MB & 43 M  & 19 M  & 21 M \\
            \brav {NanoBERT(2MB)~\cite{NanoBERT}}  & 270 K & 223 K & 1,973 MB & 163 M &	82 M & 82 M \\
            \textbf{\sname{} (ours)}           & 357 K     & 131 K     & 1,952 KB   & 272 M & 136 M & 136 M \\
            \hline \hline
            \textcolor{gray}{\emph{BERT-Tiny} (20MB)~\cite{tinyBERT}}   & \textcolor{gray}{4.4 M}     & \textcolor{gray}{786 K}     & \textcolor{gray}{20,746 MB} & \textcolor{gray}{1617 M} & \textcolor{gray}{808 M} & \textcolor{gray}{812M} \\
            \hline
        \end{tabular}
    \end{center}
    \label{table:resources}
\end{table*}

\section{Experimental setup}
\label{sec:setup}

This section outlines our experimental protocol, focused on training \sname{} and several baseline models for comparison under a strict 2~MB memory budget. 
Our comprehensive and reproducible experimental campaign spans 6 different models trained across 17 datasets, providing insights into architecture design strategies that balance performance and memory efficiency.
We detail the comparison models, training protocols, datasets, and evaluation metrics, setting the stage for the performance analysis in Section~\ref{sec:results}.

\subsection{Baseline Models and Comparisons}
\label{subs:models}

As a comparison to the proposed \sname{} model, we evaluated a diverse set of architectures, each constrained to a maximum memory footprint of 2~MB (including parameters and activations). Below, we summarize the key characteristics of these baseline models:
\begin{enumerate}
    \item BERT(2MB): 
   A scaled-down variant of the original BERT architecture \citep{BERT}, preserving the standard embedding layers and encoder blocks.
   \item MAMBA(2MB):  
   A scaled-down adaptation of the MAMBA model~\citep{MAMBA}, incorporating its native embedding mechanism and SSM-based recurrent blocks.
   \item \brav{NanoBERT(2MB):  
   A scaled-down adaptation of the NanoBERT model~\citep{NanoBERT}, based on the original BERT architecture and a novel efficient Nano Embedder.}
   \item Embedder Only:  
   This baseline leverages the Nano Embedder~\citep{NanoBERT} without any mechanism for token interaction. While not a fully functional language model, it highlights the parameter budget allocated to the embedding layer and evaluates the embedder standalone capability.
   \item Embedder + Conv:  
   Extends the \emph{Embedder Only} configuration by adding a lightweight 1D convolutional layer. This enables local token interactions within a fixed-size context window, introducing minimal parameter overhead.
\end{enumerate}

We further include BERT-Tiny~\citep{tinyBERT}, a minimal variant of BERT which is approximately $10\times$ larger than the 2~MB models evaluated in this study, as a SotA reference point. Despite its significantly larger size, it serves as a useful benchmark for performance comparison. Each model incorporates a task-specific output layer, adapted to the dataset and classification task. Given its small size and high customizability, this layer memory contribution has been excluded from the calculations of effective memory usage.

Table~\ref{table:models_params_and_memory} presents a detailed overview of the architectural parameters; in Table~\ref{table:resources} weight counts, activation sizes as well as number of sums, multiplications and memory accesses are listed for each model, including \sname{}, and illustrates their total memory footprint.

\subsection{Pre-training and Fine-tuning}
\label{subs:training_pretraining}

This section outlines the procedures used for both the pretraining and finetuning of \sname{} and the baseline models discussed in this work.

\paragraph{Pretraining Protocol}

For models supporting BERT-style pretraining~\citep{BERT}, we use the publicly available BookCorpus dataset~\citep{bookcorpus} used to train the initial GPT model by OpenAI as well as the original BERT implementation. BookCorpus comprises approximately 7,000 self-published books, totaling around 985 million words, primarily sourced from the indie eBook distribution platform Smashwords~\citep{bookcorpus}.

After training a Byte Pair Encoding (BPE) tokenizer tailored to the required dictionary size of each model, we construct sentence pairs for Masked Language Modeling (MLM) and Next-Sentence Prediction (NSP). Sentence pairs are generated by pairing half of the tokenized sentences contiguously and the other half randomly. Within each pair, tokens are masked with a 1/6 chance, and masking is applied with the following probabilities: 70\% of masked tokens are replaced with the \verb|<MASK>| token; 15\% are substituted with a random token; 15\% are left unchanged. This masking strategy promotes contextual reasoning over random guessing.

Pretraining is performed for one epoch with a batch size of 32 and a learning rate of \(5 \times 10^{-4}\) using the standard AdamW optimizer. Due to stringent memory constraints considered in this work, more complex pre-training strategies such as ELECTRA-style training~\citep{ELECTRA} had to be excluded. ELECTRA requires both a generator and a discriminator, with the generator typically being about half the size of the discriminator. Under the strict 2~MB memory constraint, it is infeasible to construct a generator of sufficient size while maintaining a capable discriminator.

In the spirit of pretraining a limited model with more data than it could ever fully absorb, we opted not to apply data augmentation or filtering. This approach aims to enhance the model generalization capabilities by exposing it to a broader range of linguistic patterns and structures inherent in the diverse BookCorpus dataset.

\paragraph{Finetuning protocol}

Following pretraining, models are finetuned on target datasets for 10 epochs using a fixed\footnote{To maintain consistency we omit complex learning rate schedulers, as different models may exhibit varying responses to specific schedules. Future work could systematically explore scheduling strategies for these models.} learning rate of \(3 \times 10^{-4}\).  
Validation is conducted at the end of each epoch using the Matthews Correlation Coefficient (MCC) for classification tasks or the Spearman Correlation Coefficient (SCC) for regression tasks. The best-performing checkpoint, determined by the highest validation metric, is saved and subsequently used for final testing.

Note that two models, namely Embedder and Embedder~+~Conv, do not undergo pretraining. Due to their extremely simple architectural structure, these models cannot effectively absorb BERT-style pretraining. Instead, they are trained directly on the target datasets for 20 epochs, doubling the standard finetuning protocol to ensure they consume the same amount of computation as the other models.

\subsection{Datasets}

\begin{table}[t]
    \caption{Datasets from the TinyNLP benchmark, covering various tasks .}
    \begin{center}
        \resizebox{\columnwidth}{!}{
        \begin{tabular}{| c | c c c |}
            \hline
            \textbf{Name} & \textbf{Size} & \textbf{Classes} & \textbf{Kind}  \\
            \hline \hline
            \href{https://huggingface.co/datasets/stanfordnlp/imdb}{\textbf{\texttt{IMDb}}}                                  & 50k   &  2    & Sentiment analysis \\ 
            \href{https://huggingface.co/datasets/fancyzhx/ag_news}{\textbf{\texttt{ag\_news}}}                              & 127.6k&  4    & Document classification \\
            \href{https://www.kaggle.com/datasets/andrewmvd/cyberbullying-classification}{\textbf{\texttt{cyberbull}}}       & 46k   &  6    & Racism classification  \\
            \href{https://huggingface.co/datasets/IBM/limit}{\textbf{\texttt{LiMiT}}}                                        & 24.6k &  2    & Movement presence \\
            \href{https://huggingface.co/datasets/dair-ai/emotion}{\textbf{\texttt{Emotion}}}                                & 20k   &  6    & Sentiment analysis \\
            \href{https://huggingface.co/datasets/xingkunliuxtracta/nlu_evaluation_data}{\textbf{\texttt{nlu}}}              & 25.7k &  18   & Request classification  \\
            \href{https://huggingface.co/datasets/benayas/snips}{\textbf{\texttt{Snips}}}                                    & 14.5k &  7    & Request classification  \\
            
            \hline
        \end{tabular}
        \label{table:tiny_datasets}
        }
    \end{center}
\end{table}

In order to meaningfully compare the performance of \sname{} and the baseline models, we use two benchmark datasets: the \emph{TinyNLP benchmark} (introduced in this paper) and GLUE~\citep{GLUE}.
In the following subsections we proceed to detail the tasks contained in both these benchmarks datasets, and the procedure for the train-validation-test splitting.

\subsubsection{The TinyNLP benchmark}
\label{sec:tinyNLP}
To better evaluate TLMs in the real-world scenarios and resource-constrained environments they are expected to operate, we introduce the \textit{TinyNLP} benchmark. 
This curated collection of existing datasets is specifically tailored to the constrained yet practical applications of language models on embedded devices.
Details of the dataset selection in the \textit{TinyNLP} benchmark are presented in Table~\ref{table:tiny_datasets}. The selection of these datasets represents application scenarios suited to models with restricted memory footprints, and is guided by the practical aim of assessing TLM deployment on embedded devices.

Building on the discussion in Section~\ref{sec:introduction}, the \textit{TinyNLP} benchmark focuses on tasks that are narrower in scope and less computationally demanding compared to standard (LLM) benchmarks. These tasks are grouped into three broad categories:
\begin{enumerate}[label=\roman*)]
    \item Request Classification: Relevant to virtual assistants in TinyML contexts, these datasets involve discerning the type of user request (e.g., requests for information, action, or assistance). As datasets focused on this kind of task, we have included \texttt{nlu}~\citep{nlu_dataset} and \texttt{Snips} in the \textit{TinyNLP} benchmark.
    \item Sentiment Analysis: Focuses on determining the emotional tone or opinion expressed in text. This commonly involves classifying content as positive, negative, or neutral, and sees wide usage in analyzing customer reviews or social media feedback. As datasets focused on this kind of task, we have included \texttt{IMDb}~\citep{imdb_dataset} and \texttt{Emotion}~\citep{emotion_dataset}.
    \item Context Understanding: Involves identifying the broader context in which text is generated. For example, distinguishing whether the text describes a particular situation or environment. As datasets focused on this kind of task, we have included \texttt{ag\_news}~\citep{ag_news_dataset}, \texttt{cyberbull}~\citep{cyberbull_dataset} and \texttt{LiMiT}~\citep{limit_dataset}.
\end{enumerate}

\begin{table}[t]
    \caption{Datasets from the \textbf{GLUE benchmark}~\cite{GLUE}, with small description about they adress.}
    \begin{center}
        \resizebox{\columnwidth}{!}{
        \begin{tabular}{| c | c c c |}
            \hline
            \textbf{Name} & \textbf{Size} & \textbf{Classes} & \textbf{Kind}  \\
            \hline \hline
            \textbf{\texttt{cola}}       & 10.7k &  2    & grammatical/semantical correctness \\ 
            \textbf{\texttt{mnli-m}}     & 403k  &  3    & correct summarization \\
            \textbf{\texttt{mnli-mm}}    & 403k  &  3    & correct summarization  \\
            \textbf{\texttt{mrpc}}       & 5.8k  &  2    & semantical equality \\
            \textbf{\texttt{qnli}}       & 116k  &  2    & question/answer entailment \\
            \textbf{\texttt{qqp}}        & 795k  &  2    & concept repetition  \\
            \textbf{\texttt{rte}}        & 5.77k &  2    & correct summarization  \\
            \textbf{\texttt{sst2}}       & 70k   &  2    & sentiment classification  \\
            \textbf{\texttt{stsb}}       & 8.63k &  /    & phrase similarity regression  \\
            \textbf{\texttt{wnli}}       & 852   &  2    & phrases entailment  \\
            
            \hline
        \end{tabular}
        \label{table:GLUE_datasets}
        }
    \end{center}
\end{table}

\subsubsection{The GLUE benchmark}
The General Language Understanding Evaluation (GLUE) benchmark~\citep{GLUE} is a widely adopted NLP benchmark comprising multiple, diverse and complex datasets, designed to test generalization and performance across diverse NLP tasks (see Table~\ref{table:GLUE_datasets}). 
It encompasses multiple subtasks, including sentiment classification and regression on sentence pairs. Because the official GLUE labels are only publicly released for the training and validation splits - and in line with prior approaches (e.g., \citep{exploring_transformer_sizes}) - we treat the validation set as our test split throughout this study.

\subsubsection{Train-Validation-Test Dataset Splittings}
For both the TinyNLP and GLUE benchmarks, each dataset is divided into training, validation, and test sets according to one of the following protocols, listed in order of priority:
\begin{enumerate}[label=\roman*)]
    \item Provided splits: When the dataset creators supply official train, validation, and test splits, we use these directly to ensure consistency with prior work.
    \item For datasets with only a single official split (e.g., train-test only), we designate the larger portion as the training set and the smaller portion as the test set. From the training set, we withhold 10\% of the samples to create a validation set.
    \item No provided splits: For datasets lacking any predefined splits, we partition the data into a 90-10 ratio for training and testing. Subsequently, 10\% of the training set is withheld to create a validation set.
\end{enumerate}

\subsection{Evaluation}
\label{sec:eval}

\begin{table*}[tbp]
    \caption{Model performance on the TinyNLP benchmark, reporting accuracy for each individual dataset and overall averages.}
    \begin{center}
        \begin{tabular}{| c | c  c  c  c  c  c  c | c |}
        \hline
        \textbf{Model}&
        \href{https://huggingface.co/datasets/stanfordnlp/imdb}{\textbf{IMDb}} &
        \href{https://huggingface.co/datasets/fancyzhx/ag_news}{\textbf{ag\_news}}  &
        \href{https://www.kaggle.com/datasets/andrewmvd/cyberbullying-classification}{\textbf{cyberbull}}  &
        \href{https://huggingface.co/datasets/IBM/limit}{\textbf{LiMiT}} &
        \href{https://huggingface.co/datasets/dair-ai/emotion}{\textbf{Emotion}} &
        \href{https://huggingface.co/datasets/xingkunliuxtracta/nlu_evaluation_data}{\textbf{nlu}} &
        \href{https://huggingface.co/datasets/benayas/snips}{\textbf{Snips}} &
        \textbf{Average} \\
        \hline
        \hline
        Embedder & 82,60 & 91,10 & 82,78 & 71,60 & 89,40 & \textbf{89,50} & \textbf{97,93} & 86,41 \\
        Embedder + conv & 84,08 & \textbf{91,50} & 83,10 & 70,32 & 89,45 & 89,33 & 97,75 & 86,50 \\
        BERT(2MB)~\cite{BERT} & 79,38 & 89,00 & 83,90 & 74,72 & 77,34 & 86,14 & 97,00 & 83,93 \\
        MAMBA(2MB)~\cite{MAMBA} & 81,86    & 89,40      & 81,38      & 74,72       & 45,72      & 70,10      & 96,40      &  77,08     \\
        \brav {NanoBERT(2MB)~\cite{NanoBERT}} &  83,32 & 90,64 & 84,06 & 74,72 & 87,20 & 86,50 & 97,90 & 86,33  \\
        \textbf{\sname{}}            & \textbf{84,10} & 90,46 & \textbf{83,97} & \textbf{76,36} & \textbf{89,58} & 88,16 & 97,67 & \textbf{87,19} \\

        \hline
        \hline
        \textcolor{gray}{\textit{BERT-Tiny}(20MB)~\cite{tinyBERT}} & \textcolor{gray}{85,69} & \textcolor{gray}{91,93} & \textcolor{gray}{83,38} & \textcolor{gray}{72,40} & \textcolor{gray}{88,86} & \textcolor{gray}{88,53} & \textcolor{gray}{98,16} & \textcolor{gray}{86,99} \\
        \hline
        \end{tabular}
        \label{table:res_tiny_comp}
    \end{center}
\end{table*}

\begin{table*}[htbp]
    \caption{Model performance on the GLUE benchmark. Metrics are MCC for CoLA, F1~score for MRPC and QQP, Spearman’s Correlation Coefficient (SCC) for STSB, and accuracy for the remaining tasks, as required for the official calculation of the overall GLUE score.}
    \begin{center}
        \resizebox{\textwidth}{!}{
        \begin{tabular}{| c | c c c c c c c c c c | c |}
        \hline
        \textbf{Model} & COLA & SST-2 & MRPC & QQP & MNLI-m & MNLI-mm & QNLI & RTE & WNLI & STSB & Score \\
        \hline \hline
        Embedder   & 9,65 & 78,90 & 62,25 & \textbf{83,28} & 62,06 & 62,17 & 65,40  & \textbf{52,73}  & 77,20  & 15,58   & 56,92    \\ 
        Embedder + conv     & 9,25  & 79,10  & 60,50  & 82,98  & 61,98  & 60,93 & 62,08  & 52,00  & 79,16  & 16,10  & 56,41    \\
        BERT(2MB) ~\cite{BERT} & -0,86 & 71,28 & 64,66 & 73,04 & 60,56 & 61,58 & 60,82 & 48,24 & 66,20 & 15,48 & 52,10\\
        MAMBA(2MB) ~\cite{MAMBA} & 2,56  & \textbf{81,16}  & 64,62  &  79,18 & 61,22  & 61,40  & 63,20  & 50,20 & 23,38  & 10,16  & 49,71  \\
        \brav {NanoBERT(2MB) ~\cite{NanoBERT}} &  9,04  & 78,82 & 65,04 & 79,96 & 63,08 & 63,30 & 63,20 & 51,76 & 87,30 & 13,46 & 57,50  \\
        \textbf{\sname{}}            & \textbf{11,01} & 79,33 & \textbf{69,19} & 83,25 & \textbf{67,83} & \textbf{68,63} & \textbf{68,92} & 49,96 & \textbf{87,61} & \textbf{49,25} & \textbf{63,50} \\
        \hline
        \hline
        \textcolor{gray}{\textit{BERT-Tiny}(20MB) ~\cite{tinyBERT}} & \textcolor{gray}{0,00}      & \textcolor{gray}{83,20}      & \textcolor{gray}{71,10}     & \textcolor{gray}{62,20}      & \textcolor{gray}{70,20}      & \textcolor{gray}{70,30}      & \textcolor{gray}{81,50}      & \textcolor{gray}{57,2}      & \textcolor{gray}{62,30}      & \textcolor{gray}{73,60}      & \textcolor{gray}{63,16}    \\
        \hline
        \end{tabular}

    \label{table:res_glue_comp1}
    }
    \end{center}

\end{table*}

The pretraining of all models on the BookCorpus dataset is conducted once, while the fine-tuning phase on each target dataset is repeated five times with different random seeds corresponding to different AdamW mini-batch shuffling, to ensure robustness of the results. Evaluation metrics are computed as the average of these five runs.

For the sake of simplicity, in the experimental results reported in Sec.~\ref{sec:results}, as evaluation metrics we focus on Accuracy for the TinyNLP benchmark, and on the metric used for computing the average Score in each dataset in the GLUE benchmark: SCC for STSB, MCC for CoLA, F1~score for QQP and MRPC, Accuracy for the remaining GLUE tasks. 



\section{Experimental Results}
\label{sec:results}

We evaluate \sname{} and the baseline models on the TinyNLP and GLUE benchmark datasets, reporting the obtained results respectively in Tables~\ref{table:res_tiny_comp}~and~\ref{table:res_glue_comp1}.
Average results for the two benchmark datasets were also calculated and are reported in the last column of the Tables.


On the TinyNLP benchmark, \sname{} demonstrates the best performance compared to the other models, achieving an Average Accuracy of $87.19\%$. 
Notably, \sname{} outperforms BERT-Tiny, which requires around $10\times$ more memory but only achieves the second-highest average Accuracy of $86.99\%$. Interestingly, the only models in the 2~MB range that offer comparable results were the Embedder and Embedder~+~Conv configurations. Despite their seemingly simplistic design, these models perform well on the TinyNLP tasks. \brav{The optimized NanoBERT model achieved an average accuracy of $86,33\%$, falling slightly behind the Embedder and Embedder~+~Conv models.} 
These results highlight the Embedder models' ability to handle lightweight tasks effectively. The 2~MB down-scaled versions of the BERT and MAMBA models, on the other hand, scored lower on Average Accuracy, indicating that these models may be less suitable for environments with stringent memory budgets. This suggests that the overall architecture of \sname{}, with highly optimized embedding and attention structures, is particularly well-suited for the TinyNLP classification tasks in memory-constrained scenarios, with respect to down-scaled versions of standard models. 

On the GLUE benchmark, \sname{} emerged as the top-performing model, achieving an average score of $63.50$, outperforming again BERT-Tiny, which achieved the second-highest average Accuracy of $63.16$. All the other models within the 2~MB budget achieved significantly lower performance, and, also in this case, the Embedder and Embedder~+~Conv models (with an average score of $56.92$ and $56.41$, respectively) performed better than the BERT and MAMBA models ($52.10$ and $49.71$) by a large margin. \brav{In this benchmark, the NanoBERT model, with an average score of $57.50$ performed slightly better than the Embedder and Embedder~+~Conv models, but still many points behind \sname{} and BERT-Tiny.}





\subsection{Discussion of the Results}

In the TinyNLP benchmark, the model performs particularly well on sentiment and emotion analysis (IMDb, Emotion) and intent recognition tasks (Snips, nlu), indicating a strong ability to capture semantic and affective cues. The good performance on LiMiT also suggests improved handling of syntactic and structural relations, which are typically challenging for lightweight architectures. Conversely, results on document classification tasks (e.g., AG~News) show smaller gains, likely due to the reduced capacity to model long-range dependencies.

\begin{table*}[tbp]
    \caption{Scaled Models' memory and computational requirements. Columns represent weights, activations, total size, number of additions (sums), multiplications (mults), and memory accesses.}
    \begin{center}
        \begin{tabular}{|c|c c c |c c c|}
            \hline
            \textbf{Model} & \textbf{Params} & \textbf{Activ}. & \textbf{Total (MB)} & \textbf{Mem. accesses} & \textbf{Sums} & \textbf{Mults}  \\
            \hline \hline
            \textbf{EmbBERT-Nano}  & 64K  & 98K   & 648 KB & 52 M & 26 M & 26 M \\
            \textbf{EmbBERT-Micro} & 179K & 131K  & 1.24 MB & 137 M & 69 M & 69 M \\
            \textbf{EmbBERT}       & 357K & 131K  & 1.95 MB & 272 M & 136 M & 136 M \\
            \textbf{EmbBERT-Med}   & 2M   & 328K  & 9.6 MB & 2 B & 812 M & 812 M \\
            \textbf{EmbBERT-Big}   & 7M   & 3M    & 39.71 MB &32 B & 16 B & 16 B \\
            \hline
        \end{tabular}
    \end{center}
    \label{table:models_params_scaling}
\end{table*}

In the GLUE benchmark, \sname{} shows clear advantages on syntactic understanding and semantic similarity tasks. The gains on MRPC and QQP further indicate effective pairwise sentence encoding, essential for paraphrase and entailment recognition. Nevertheless, \sname{} shows weaker results on RTE, a task requiring more complex logical reasoning and world knowledge - an expected limitation for models of this scale. 

Thus the results from both the TinyNLP and GLUE benchmarks establish the proposed \sname{} as the current SotA Language Model for TinyML hardwares and NLP applications. Among the available LMs, it obtains the best experimentally observed balance between memory requirements and task performance. The experiments demonstrated also that within the strict constraints of a 2~MB memory budget, standard LM architectures such as BERT and MAMBA struggle to reach even comparable results to very simple models (Embedder and Embedder~+~Conv)\brav{, while scaled-down architectures designed for efficiency, such as NanoBERT, can perform slightly better than the simple models, but still worse than the proposed EmbBert}.

\section{Quantizing and Scaling \sname{}}
\label{sec:quant_n_scale}

In this section, we explore the robustness of the model to 8-bit post-training quantization, and we test the scalability of the \sname{} architecture, exploring configurations ranging from $0.25\times$ to $20\times$ the memory requirements of the base configuration.

\subsection{Quantizing \sname{}} 
\label{subs:quantization}

Quantization is a critical step in deploying deep learning models on embedded devices. It reduces memory requirements, enables execution on hardware that lacks support for 32- or 64-bit precision computations, and can significantly improve inference speed on devices that support lower-precision operations \cite{shalbyDynamicQuantization,MixedPrecisionQuantization,nnquantization}. For this reason, we evaluated the resilience of the proposed \sname{} under a quantization scheme. 

Specifically, we applied an 8-bit block-wise post-training quantization approach implemented in the ~\citep{bitsandbytes} higgingface library. This method applies per-block quantization to model weights, where each block (of 64 parameters) is scaled independently based on its local dynamic range. The resulting 8-bit representation significantly reduces memory usage while preserving numerical precision. All computations are performed in mixed precision, with weights quantized to 8-bit and activations maintained in FP16 to ensure stability and performance.
Further parameter-efficient fine-tuning (PEFT) is performed for additional two epochs with the 8-bit AdamW optimizer and a fixed learning rate of $1\times 10^{-4}$. By doing so, we modify only a small subset of parameters (approximately 8\% of the total weights), allowing the model to better adapt to specific tasks while maintaining the advantages of reduced precision.

The complete results of the application of the quantization scheme to the \sname{} model is reported in \ref{appendix:completeresults}. The results show that the quantized version of \sname{} has minimal or no performance degradation with respect to the full precision model. In particular, the quantized version of \sname{} showed a marginal improvement on the TinyNLP benchmark, obtaining an average accuracy of $88.17\%$; while on the GLUE benchmark, \sname{} achieved an overall GLUE score of $62.81$, demonstrating exceptional robustness to quantization. This represents a minimal performance drop of $-0.7$ percentage points compared to the unquantized \sname{} version.

The quantization process gives substantial memory savings, reducing the total memory required to store and execute \sname{} from the around 2~MB of the unquantized \sname{} to just 781~kB, considering both weights and activations (a $2.4\times$ reduction in memory demand). 


\subsection{Scaling the \sname{} architecture}
\label{subsec:scaling}

\begin{figure}[t]
    \centering
    \includegraphics[width=\linewidth]{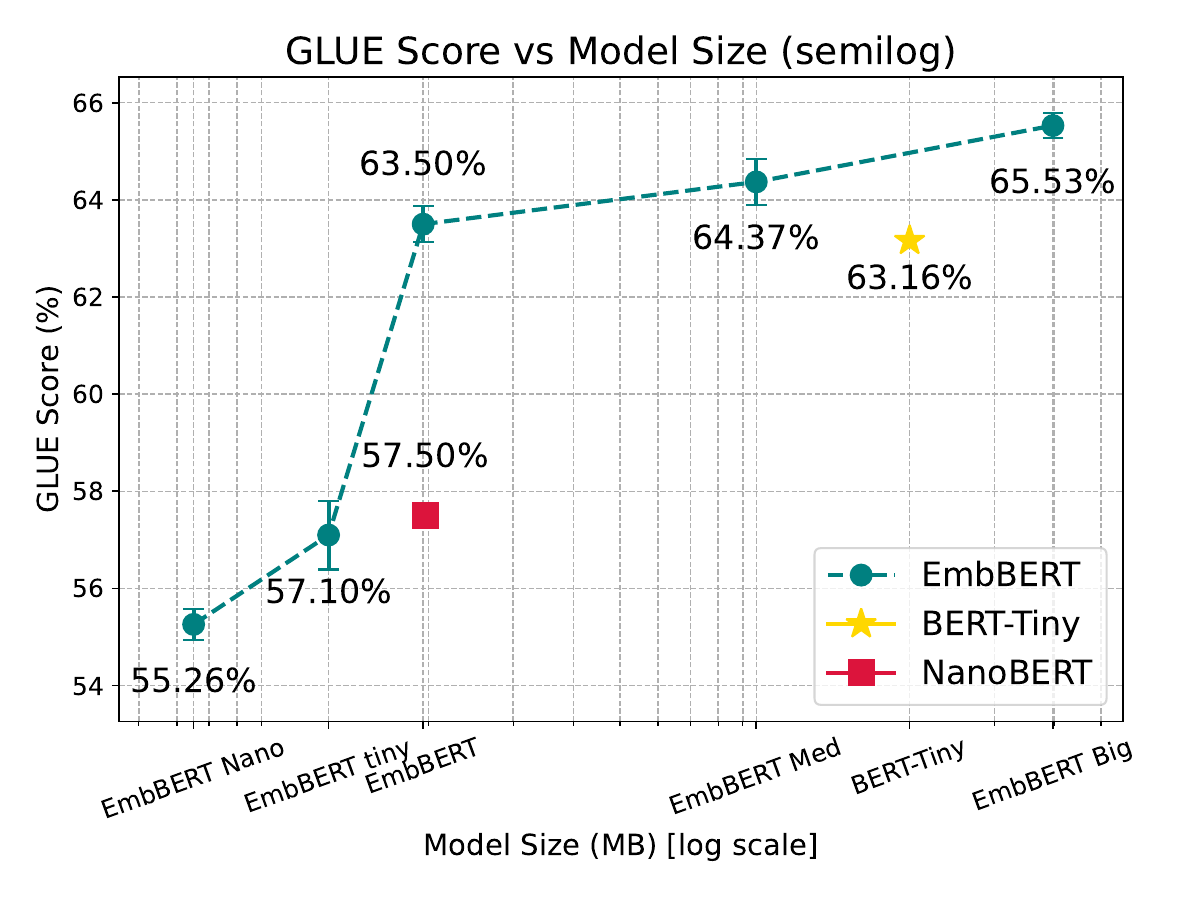} 
    \caption{GLUE scores of the different scaled models are represented in comparison with the SotA BERT-Tiny model and the retrained NanoBERT (2MB). On the x axis in logarithmic scale we have the model sizes and on the y axis their relative GLUE scores are represented} 
    \label{fig:size-vs-score}
\end{figure}

To gain deeper insights into the trade-off between model compactness and downstream performance, we analyzed four different scaled configurations of \sname. These additional versions of the architecture---\sname-Nano (0.5~MB), \sname-Tiny (1~MB), \sname-Med (10~MB), and \sname-Big (40~MB)--- are compared to both the original\sname{} model as well as to the smallest SotA baseline, BERT-Tiny (20~MB). Details on the memory and computational requirements are reported in Table \ref{table:models_params_scaling}, while we provide details on the used hyperparameters and how we selected them for this solution in \ref{appendix:param_scaling}. All models were pretrained on the book corpus and then fine-tuned exclusively on the GLUE benchmark.
The obtained results in terms of score are reported in Figure~\ref{fig:size-vs-score} along with the memory occupations of the models and BERT-Tiny as reference, while we report the complete results in \ref{appendix:completeresults}. 
The average score grows monotonically with model size, with a huge gap between the \sname{}-tiny version and \sname{}. Extreme compression yields functional models that are extremely efficient, but at the cost of a performance gap on tasks requiring nuanced reasoning.
Intermediate compression, such as with \sname{} and \sname-Med, offers the most balanced trade-off, achieving near-BERT-Tiny performance with a $2$ to $10$ times smaller footprint.
\sname{}-Big demonstrates that the architecture can scale effectively: at just double the size of BERT-Tiny, \sname-Big surpasses it consistently, proving that the architectural modifications transfer well to higher capacity regimes.

These findings suggest that \sname{} offers a flexible scaling curve: \brav{while \sname{} was designed specifically for the 2MB memory budget, the core ideas in its designs can be used in multiple settings and with different memory budgets,} from ultra-compact sub-megabyte models suitable for resource-constrained devices, to larger yet still efficient models that outperform comparably sized baselines. \brav{Although the scaling reported in Fig. \ref{fig:size-vs-score} looks promising, the accuracy of the EmbBERT architecture at higher memory budgets remains to be tested, although it appears unlikely that it could outperform models specifically designed for that regime.}  Overall, \sname{} demonstrates both scalability and adaptability, reinforcing its potential as a practical language model family for resource-constrained environments.

\begin{figure*}[t]
    \centering
    \begin{minipage}[t]{0.48\textwidth} 
        \centering
        \includegraphics[width=\textwidth]{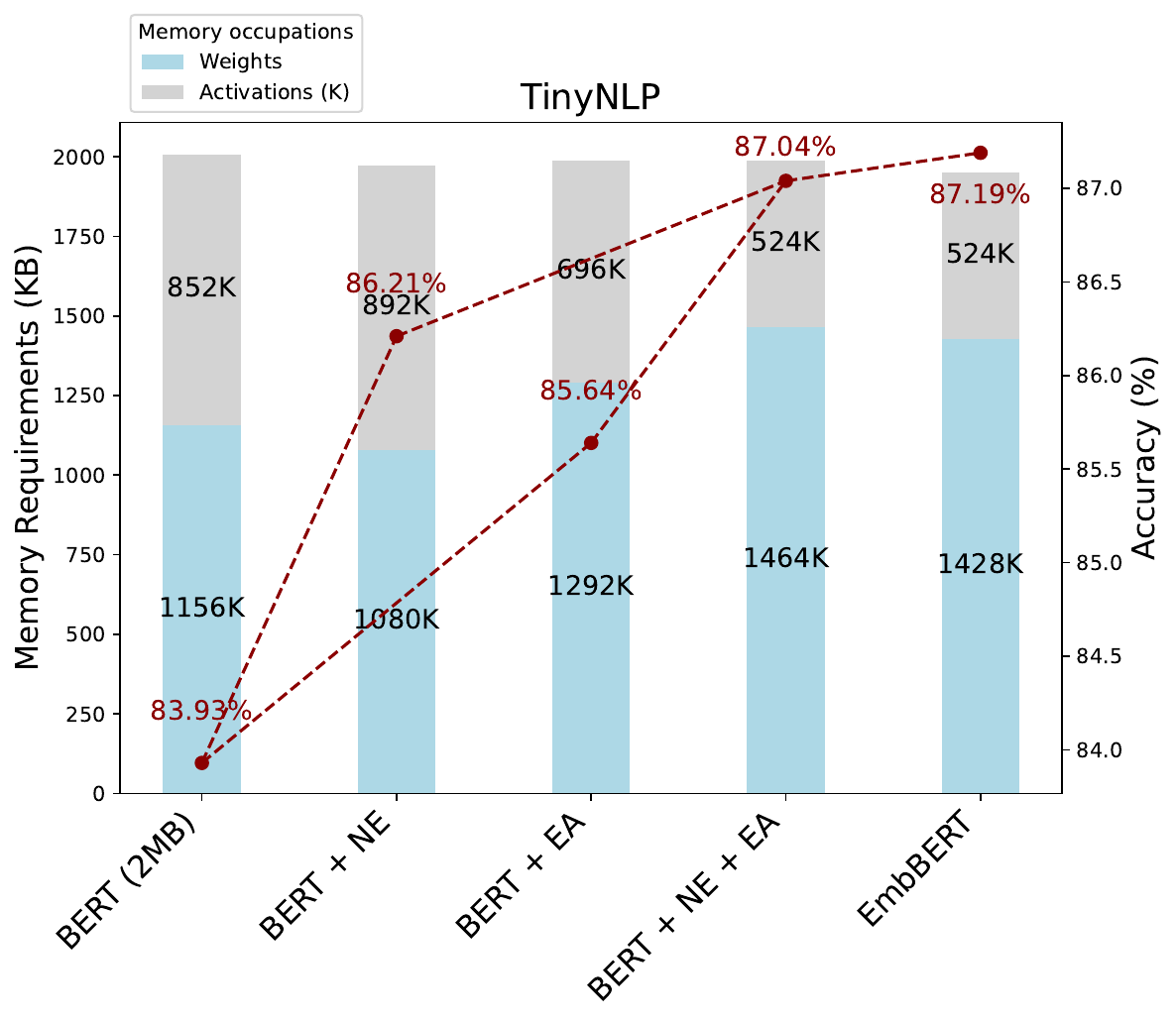} %
    \end{minipage}%
    \hfill
    \begin{minipage}[t]{0.48\textwidth}
        \centering
        \includegraphics[width=\textwidth]{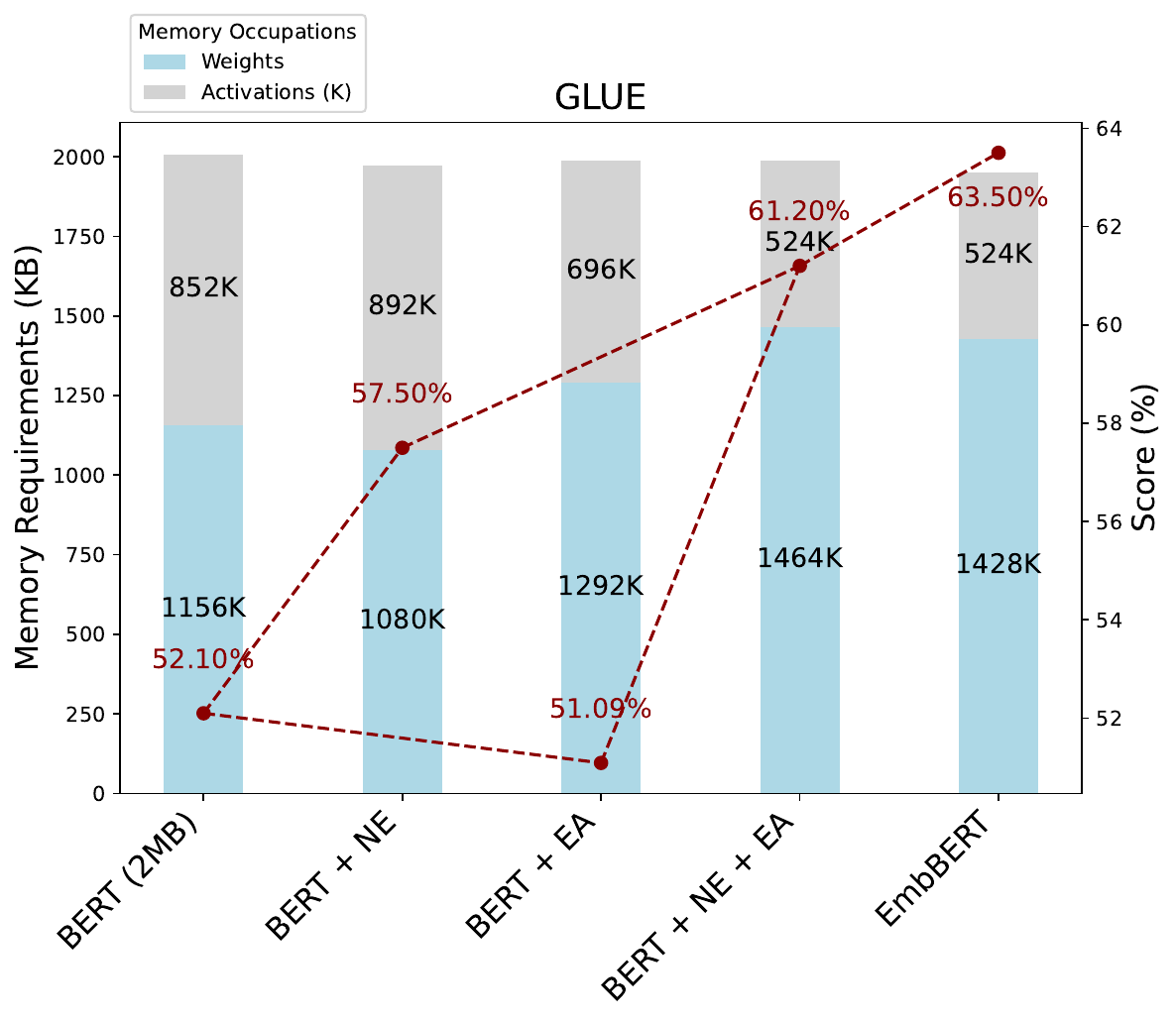} %
    \end{minipage}
    \caption{Memory occupation for weights (blue) and activations (grey), and Accuracy/Score results obtained by each model analyzed in the ablation study on the TinyNLP benchmark (left panel) and on the GLUE benchmark (right panel). }
    \label{Fig:memory_ablation}
\end{figure*}


\section{Ablation Study}
\label{sec:ablation}

This ablation study aims to delve deeper into the contribution of each of \sname{} individual architectural components. Furthermore, we analyze how much each of the models tested in the Experimental Results section benefit from the application of a pre-training procedudre.

\subsection{Evaluating the impact of the architectural components of \sname{}}

In this section, we analyze the contributions of \sname{} architectural components to its overall performance on both the TinyNLP and GLUE benchmarks. 
Taking BERT (2~MB) as a baseline, we systematically introduced improved and optimized key components, defining in this way \sname{}. Through this process, we evaluate the individual and collective impact of these changes, always adhering to the strict memory constraints of 2~MB.
Figure~\ref{Fig:memory_ablation} shows the memory requirements of various model configurations alongside their corresponding accuracy on TinyNLP and score on GLUE. Complete results can be on the 2 benchmarks can be found in \ref{appendix:completeresults}.

\paragraph{Base model - BERT(2~MB)}
BERT(2~MB) is a compressed variant of BERT that serves as our vanilla baseline model. Compared to the larger BERT-Tiny model~\citep{tinyBERT}, which has a $10\times$ memory footprint, BERT(2~MB) shows a notable performance degradation that could be attributed to its reduced parameter count.
Despite these limitations, BERT(2~MB) marks a first step toward adapting transformer architectures for ultra-low-memory environments, demonstrating the feasibility of scaling down LM models while maintaining some level of task performance.

\paragraph{BERT + Nano Embedder (BERT + NE)}
To address the limitations of BERT(2~MB), we replace its Embedder with the Nano Embedder~\citep{NanoBERT}, which is designed to optimize embedding representations without increasing the overall parameter count. This substitution expands the effective vocabulary space within the same memory budget, resulting in notable performance improvements on both the TinyNLP and GLUE benchmarks.

\paragraph{BERT + Efficient Attention (BERT + EA)}
To reduce activation overhead, we proceed to replace the default multi-head attention module with Efficient Attention~\citep{efficient_attention}, aiming to lower weight and activation memory costs. This reduction allows for an increased embedding dimensionality and/or additional layers. This modification significantly improves performance on the TinyNLP benchmark but, when not paired with other architectural modules, results in a slight decrease in performance with respect to the base BERT(2~MB) model on the GLUE benchmark.

\paragraph{BERT + NE + EA} 
We combine the Nano Embedder with Efficient Attention to create the BERT~+~NE~+~EA model, leveraging a broader vocabulary together with reduced weights and activations overhead. 
This combination leads to a performance gain on both TinyNLP and GLUE tasks, where BERT~+~NE~+~EA achieves respectively an accuracy of $87.04\%$ and a score of $61.20$, i.e. respectively over $+3$ and $+9$ points compared to the original BERT(2~MB).
These results highlight the advantage of combining embedding efficiency with an optimized attention mechanism in ultra-compact models.

\paragraph{\sname}
Finally, by integrating a parallel path consisting of a 1D convolution followed by a fully connected layer, and merging its output with the efficient attention branch through a weighted sum, we obtain the \sname{} architecture. This addition introduces a lightweight feed-forward component to the main attention path of BERT~+~NE~+~EA, incurring minimal memory overhead. 
Despite its simplicity, this modification yields a modest improvement on the TinyNLP benchmark and achieves a substantial performance gain—over $+2$ points in the GLUE score—compared to the BERT~+~NE~+~EA model.

\vspace{5mm}

Through the comprehensive ablation study performed in this section, we have examined the contributions of key architectural components. Maintaining the total memory usage below the 2~MB budget throughout the study, we have demonstrated that the inclusion of these architectural components in \sname{} leads to an Average Accuracy improvement of $+3.26$ percentage points on the TinyNLP benchmark and a $+11.40$ point increase on the GLUE benchmark score, with respect to the original BERT(2MB) model.

\begin{figure}[t]
    \centering
    \includegraphics[width=\linewidth]{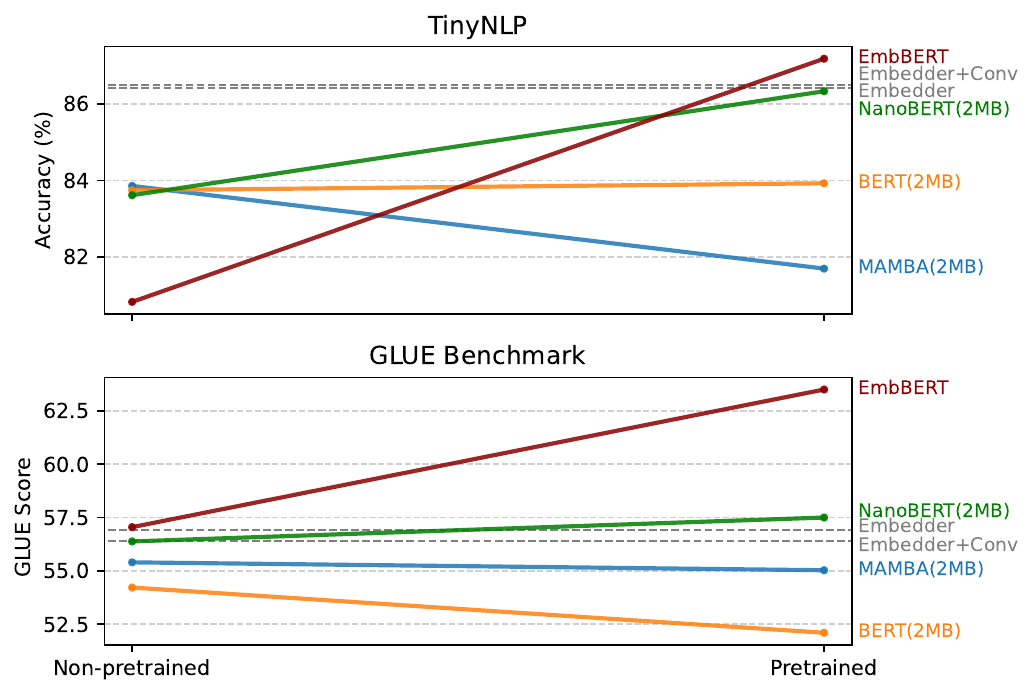} 
    \caption{Average Accuracy and GLUE scores of the Non-pretrained and Pretrained versions of the models on the TinyNLP and GLUE benchmarks.}
    \label{fig:pretrain_abl}
\end{figure}

\subsection{Evaluating the effect of pretraining on the models}

We evaluated the results of the non-trivial language models (i.e., BERT (2MB), MAMBA (2MB), NanoBERT(2MB), and \sname{}) both with and without pretraining on the book corpus, and reported the average results in Figure \ref{fig:pretrain_abl}. Complete results of both pretrained and non-pretrained models can be found in \ref{appendix:completeresults}. In the figure, it is visible how \sname{} \brav{and NanoBERT(2MB) are} the only models within the 2MB budget that can effectively make use of the pretraining: Both the pretrained versions of BERT (2MB) and MAMBA (2MB), in fact, in both the benchmarks, obtain performance which are equivalent, or even worse of the ones of their respective non-pretrained counterpart. The average accuracy of \sname{}  jumps from $80.82$ to $87.19$ in the TinyNLP benchmark, and the GLUE score from $57.05$ to $63.50$ in the GLUE benchmarks, an improvement of roughly $6.5$ percentage points in both benchmarks. \brav{Conversely, while NanoBERT (2MB) can obtain some improvements using pretraining, they are much smaller: roughly $3$ and $1$ percentage points in the TinyNLP and GLUE benchmark, respectively.}




\section{Conclusions}
\label{sec:conclusions}



In this work, we presented \sname{}, a novel language model specifically designed for tiny devices and extreme resource constraints. \sname{} achieves Accuracies and Scores comparable to the ones of models with up to $10\times$ its memory footprint, and stands out as the highest-performing solution within the 2~MB memory budget explored in this paper. 


By leveraging an innovative architectural design specifically tailored for extremely memory-constrained environments, \sname{} effectively balances parameter efficiency and competitive accuracy. Its architectural redesign significantly reduces the model’s memory footprint while maintaining high performance both on the newly proposed TinyNLP benchmark and on the GLUE benchmark, crucially demonstrating the effectiveness and the efficiency of \sname{} in resource-constrained environments. The model demonstrated resilience to 8-bit quantization, and its architecture showed an interesting scalability potential within the memory range of embedded devices.

Future research will explore further compression techniques, novel architectural designs, targeted knowledge distillation, mixed-precision quantization, and even more extreme quantizations tailored to emerging hardware accelerators (e.g., 1-bit quantization). Combining these advancements with next-generation hardware has the potential to further optimize model memory and computation footprints while preserving, or even enhancing, performance. Our work establishes a systematic foundation for designing efficient language models capable of operating effectively within the most stringent memory constraints.

\section*{Acknowledgment}

This paper is supported by PNRR-PE-AI FAIR project
funded by the NextGeneration EU program.

\bibliographystyle{elsarticle-num} 
\bibliography{main.bib} 

\appendix

\section{Exact Computation of Memory and Computational Cost of LLM Layers}
\label{appendix:memory}

The architecture of Large Language Models (LLMs)
is primarily based on the Transformer model introduced by~\citep{transformers}. This architecture has revolutionized natural language processing by enabling models to effectively handle long-range dependencies in text. Encoder-based text classification models typically consist of two main components: an embedder and an encoder. The encoder, in turn, is primarily composed of an Attention Mechanism and a Feed-Forward Neural Network.

In this section, we provide a comprehensive analytical, layer-by-layer evaluation of the memory and computational requirements of common components used in Language Models, as well as an overall view of their functionality.
We emphasize that our computational evaluations account for the complexity and memory access profiling associated with all needed model layers. Particular attention is given to CPU-based operations, including summation, multiplication, and memory retrievals.
For memory evaluation, we assume a non-parallel program that retains only the minimum required data in memory to execute effectively. This approach reflects realistic constraints in resource-constrained environments, such as those encountered in TinyML applications.

The following assumptions are made during our detailed computation of weight and activation matrices memory requirements for hardware deployment:
\begin{itemize} 
    \item Operations such as sums, element-wise multiplication, and activation functions are performed in-place, occupying only the memory of the input matrices, as intermediate results are discarded.
    \item Matrix multiplications require memory for both input matrices as well as the output matrix.
    \item Fully connected layers are treated as matrix multiplications where only one input and the output matrix contribute to activation memory, since weights do not increase activation memory requirements.
    \item The maximum memory consumption per layer is recorded at peak usage during processing.
    \item All calculations are based on inference-only processing, without accounting for training-related overheads.
\end{itemize}
The final memory occupations, memory accesses, and formulas for sums and multiplications for each block are provided in Tables~\ref{table:w&a_noMAMBA_full}~and~\ref{table:complexity_appendix}.

\subsection{BERT}
BERT (Bidirectional Encoder Representations from Transformers)~\citep{BERT} is a Transformer Encoder-based foundational NLP model widely used for tasks such as text classification, question answering, and text generation. It leverages Transformer layers to generate contextualized representations of input text, capturing both left-to-right and right-to-left dependencies.
The architecture of BERT typically consists of an Embedder, followed by a series of Attention and Feed-Forward layers, interleaved with normalization layers. These components are repeated $N$ times.

\begin{figure}[t]
    \centerline{
        \includegraphics[height=.30\textheight,keepaspectratio]{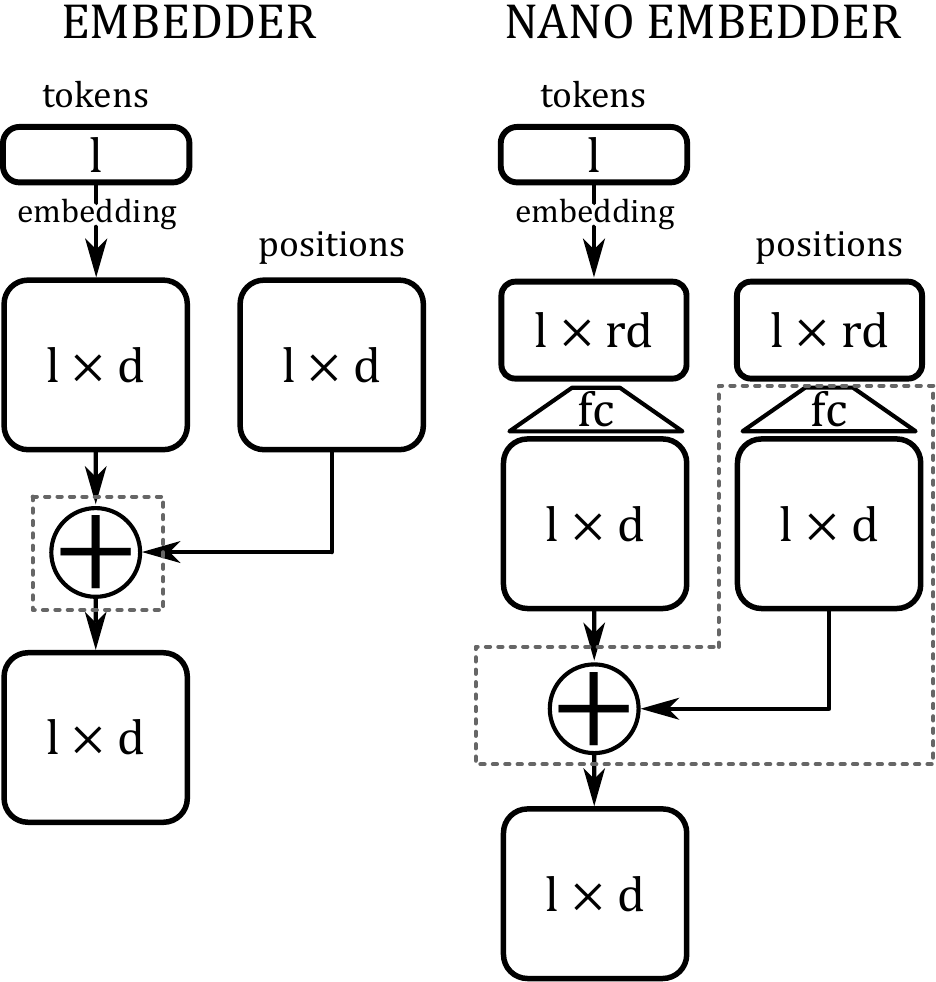}
    }
    \caption{Memory layout of the Standard Embedder (left) and Nano Embedder (right) layers. $\ell$ is the sentence length, $d$ is the embedding dimension, and the trapezoid labeled~\textit{fc} denotes fully connected layers. The dashed gray box highlights the operations requiring the maximum activation size. The Nano Embedder reduces the number of weights while maintaining a similar activation size.}
    \label{fig:embedders_memory}
\end{figure}

\paragraph{Embedder}
The standard Embedder, illustrated in Fig.~\ref{fig:embedders_memory}, is responsible for generating token embeddings, positional encodings, and segment embeddings using learned dictionaries. These embeddings are then summed to produce the final input encoding fed into the model.

The total parameter count for the embedder, $W_{emb}$, is calculated as the sum of the sizes of the token embedding matrix ($v \cdot d$), the positional embedding matrix ($\ell \cdot d$), and segment embedding matrix ($2d$):
\begin{equation}
    W_{emb} = d \cdot (v + \ell + 2),
\end{equation}
where $v$ is the vocabulary size, $\ell$ is the sequence length, and~$d$ is the embedding dimension.

The maximum activation size, $A_{emb}$, results from storing token, positional, and segment embeddings as matrices of size $\ell \cdot d$ during inference. These embeddings are summed in pairs, leading to: 
\begin{equation}
    A_{emb} = 2d \cdot \ell.
\end{equation}

The embedding operations required to compute the output of this layer involve $\ell \cdot (4d + 2) + 2d$ memory accesses and $\ell \cdot 2d$ summations.

\paragraph{Attention}

The standard Attention Mechanism~\citep{transformers} allows models to selectively focus on the most relevant parts of an input sequence.
Initially designed for machine translation, attention assigns varying "weights" to tokens based on their relevance to the task or context, enabling models to capture dependencies between distant words.

The self-attention variant computes relationships within a sequence by enabling each token to attend to all others, creating contextualized representations that encode both local and global dependencies. The input is processed through three fully connected layers to produce the Query, Key, and Value matrices, each with size \(d^2\). 
This step generates an activation size of \(4 \cdot d \cdot \ell\) and requires: i) \(6\ell \cdot d^2\) memory accesses, ii) \(3\ell \cdot d^2\) summations, and iii) \(3\ell \cdot d^2\) multiplications.

The Query and Key matrices are then multiplied to compute a Weight matrix of size \(\ell \times \ell\) for each of the \(h\) attention heads.
Due to the quadratic growth in the Weight matrix, the context length has a significant impact on activation size, which increases to \(3 d \cdot \ell + \ell^2 \cdot h\). This step also adds: i) \(2\ell^2 \cdot d \cdot h\) memory accesses, ii) \(\ell^2 \cdot d \cdot h\) summations, and iii) \(\ell^2 \cdot d \cdot h\) multiplications.

The Weight matrix undergoes a softmax operation, contributing: i) \(2\ell^2 \cdot h\) memory accesses, ii) \(\ell^2 \cdot h\) summations, and iii) \(3\ell^2 \cdot h\) multiplications,
without increasing activation size. 

Next, the Weighted matrix is multiplied by the Value matrix, producing an output activation size of \(2 d \cdot \ell + \ell^2 \cdot h\). This step introduces i) \(2\ell^2 \cdot d \cdot h\) memory accesses, ii) \(\ell^2 \cdot d \cdot h\) summations, and iii) \(\ell^2 \cdot d \cdot h\) multiplications.

Finally, the output passes through a fully connected layer of size \(d^2 + d\) with a skip connection. Although the activation size remains unchanged, this stage involves: i) \(\ell \cdot d^2 \cdot 2\) memory accesses, ii) \(\ell \cdot d^2 + \ell \cdot d\) summations, and iii) \(\ell \cdot d^2\) multiplications.

\begin{figure}[t]
    \centerline{
        \includegraphics[height=.5\textheight,keepaspectratio]{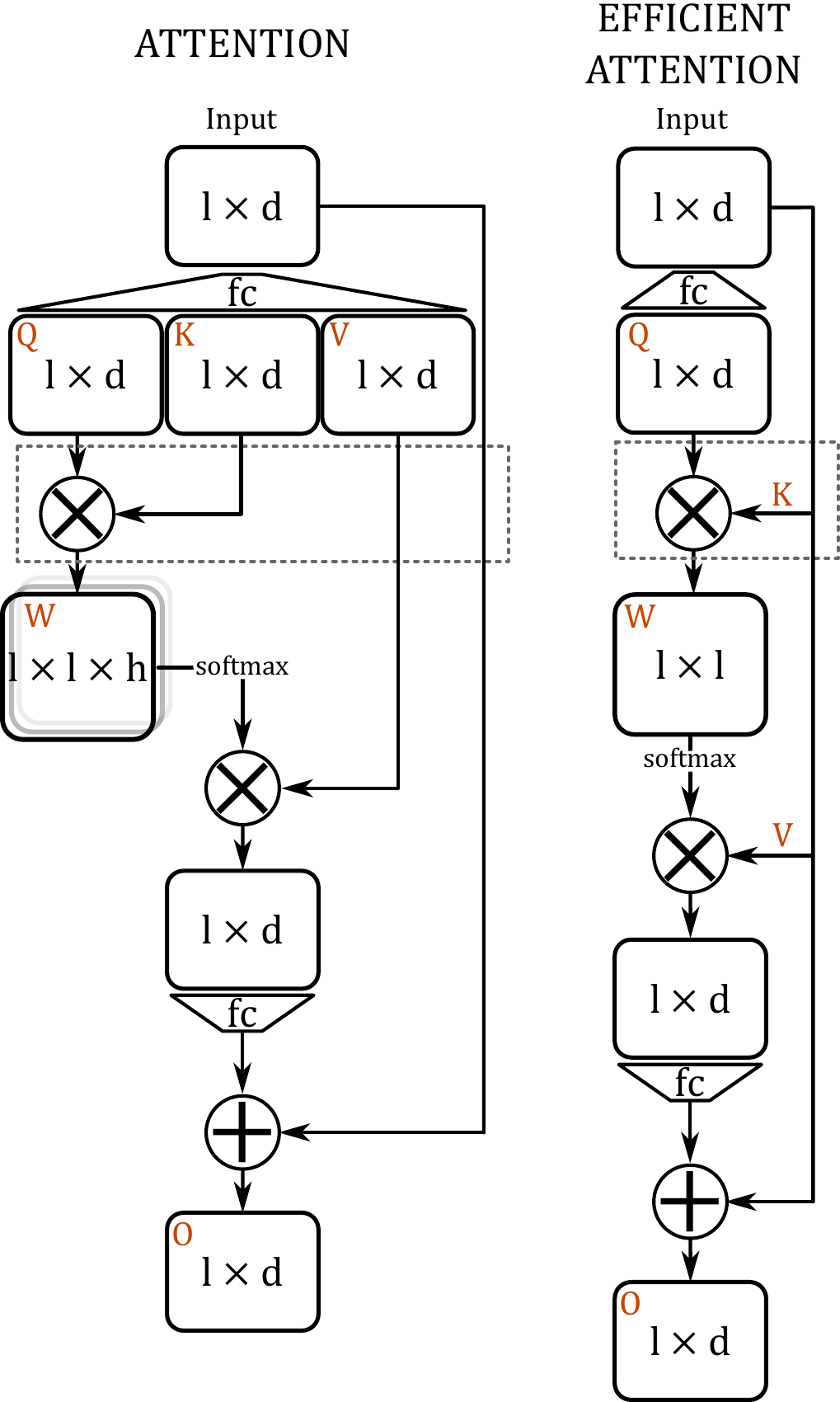}
    }
    \caption{Memory layout of the Attention (left) and Efficient Attention (right) layers. $\ell$ is the sentence length, $d$ is the embedding dimension, and the trapezoid labeled~\textit{fc} denotes fully connected layers. The dashed gray box highlights the operations requiring the maximum activation size. Efficient Attention significantly reduces activation size.}
    \label{fig:attentions_memory}
\end{figure}

\subsection{Nano Embedder}

The NanoBERT model, introduced in~\citep{NanoBERT}, is an ultra-compact version of BERT, designed for efficient operation on resource-constrained devices. 
It achieves this by employing techniques such as word embedding factorization and Low-Rank Adaptation (LoRA). 
A key component of its efficiency is the Nano Embedder (shown in Fig.~\ref{fig:embedders_memory}), which serves the same purpose as the standard Embedder but introduces a critical optimization: instead of embedding tokens and positions directly into vectors of size~$d$, it maps these inputs to a reduced embedding dimension~$r_d$ using a fully connected layer. 
This reduced embedding is then projected back to the original dimension~$d$ through a fully connected layer. Segment embeddings are excluded from this process.

This approach modifies the parameter count to:
\begin{equation}
    W_{nemb} = r_d \cdot (v + \ell + 2 d) + 2 d,
\end{equation}
which can be lower than that of the Standard Embedder if  $r_d$ is sufficiently small relative to $d$. 
However, the total activation size, $A_{nemb}$, increases slightly due to the projection step, resulting in:
\begin{equation}
    A_{nemb} = r_d \cdot \ell + 2 d \cdot \ell.
\end{equation}

During inference, the Nano Embedder performs the following operations: i) $\ell + 2\ell \cdot r_d$ memory accesses for token and position encoding, followed by ii) $4\ell \cdot r_d \cdot d$ memory accesses, iii) $2\ell \cdot r_d \cdot d + \ell \cdot d$ summations, and iv) $\ell \cdot r_d \cdot d \cdot 2$ multiplications for the linear layers; and if segment embeddings are needed another \(\ell + d \cdot + \ell \cdot d \) memory accesses and \(\ell \cdot d\) summations.

This balance of parameter efficiency with a slight increase in activation memory illustrates the Nano Embedder ability to reduce the overall model size while maintaining embedding functionality. It is particularly advantageous in resource-constrained scenarios or when prioritizing higher dictionary sizes for improved performance.

\subsection{Efficient Attention}
In~\citep{efficient_attention}, the authors introduce Efficient Attention, an approach to effectively halve the parameter count of the Standard Attention mechanism. Instead of using three fully connected layers at the beginning of the layer and one at the output, Efficient Attention employs only one fully connected layer to generate the Query matrix. The Key and Value matrices are directly taken as the input. A single fully connected layer processes the attention output - both fully connected layers have dimensions~\(d^2\).

The total number of parameters, $W_{eatt}$, for the Efficient Attention layers is calculated as:
\begin{equation}
    W_{eatt} = 2 d ^ 2
\end{equation}

Like the Standard Attention layer, the Efficient Attention layer’s highest activation size occurs during the matrix multiplication step. This step produces an activation size of:
\begin{equation}
    A_{eatt} = 2\ell \cdot d + \ell^2
\end{equation}

To calculate the operations and memory accesses required, the expressions from the Standard Attention layer can be simplified by omitting terms associated with the two additional fully connected layers, as well as the $h$ (attention heads) term from all equations (see Table~\ref{table:complexity_appendix}). 
As shown in \citep{efficient_attention}, these modifications are such that not hinder the effective modeling capabilities of the attention layer. It retains similar contextual performance while significantly reducing computational cost, making it highly advantageous for resource-constrained scenarios.

\begin{table*}[t]
    \caption{Formulas for calculating weights and activation sizes per layer, based on their architectural parameters.}
    \begin{center}
        \begin{tabular}{|c | c c|}
        \hline
        \textbf{Layers} & \textbf{Weights} & \textbf{Activations} \\
        \hline \hline
        \textbf{Embedder}               & $(v + \ell +2)\cdot d$ 
                                        & $2 d\cdot \ell$ \\
        \textbf{NanoEmbedder}           & \(r_d \cdot (v + \ell + 2 d ) + 2 d\)
                                        & \(r_d \cdot \ell + 2 d \cdot \ell\) \\
        \textbf{Normalization}          & \(2 d\)
                                        & \(2 d \cdot \ell\) \\
        \textbf{Feed Forward}           & \(2 d^2 \cdot \alpha \) 
                                        & \(d \cdot \ell \cdot ( 2 \alpha )\)\\
        \textbf{Attention}              & \(4 d^2 \)
                                        & \(4 d \cdot \ell + \ell^2 \cdot h\)\\
        \textbf{Efficient Attention}    & \(2 d^2 \) 
                                        & \(2 d \cdot \ell + \ell^2 \)\\
        \textbf{Eff Attention + Conv Skip} & \(2 d^2 + d^2 \cdot \alpha + c \cdot d \alpha\)
                                        & \(\max ( 2 d \cdot \ell + \ell^2 ; \  d \cdot \ell (2 + \alpha )) \)\\
        \hline
        \end{tabular}
    \end{center}
    \label{table:w&a_noMAMBA_full}
\end{table*}

\begin{table*}[ht]
    \caption{Formulas for calculating memory accesses, summations and multiplication performed by each layer, based on their architectural parameters.}
    \centering
    \resizebox{\textwidth}{!}{
        \begin{tabular}{|c | c c c|}
        \hline
        \textbf{Layers} & \textbf{Memory accesses} & \textbf{Summations} & \textbf{Multiplications} \\
        \hline \hline
                        
        \textbf{Embedder}               & \(\ell \cdot (4d + 2) + 2d\)  
                                        & \(\ell \cdot 2d\) 
                                        & 0   \\
        
        \textbf{NanoEmbedder}           & \(\ell \cdot (r_d \cdot 4d + d + 2r_d + 2) + 2d\)    
                                        & \(2 \ell \cdot d \cdot (r_d + 1)\)
                                        & \(\ell \cdot r_d \cdot d \cdot 2\) \\ 

        \textbf{Normalization}          & \((\ell+1) \cdot d \cdot 2\)    
                                        & \(\ell \cdot d\)
                                        & \(\ell \cdot d\)\\
                                        
        \textbf{Feed Forward}           & \(4 \ell \cdot d \cdot (d \cdot \alpha + 1)\)
                                        & \(2 \ell \cdot d \cdot (d \cdot \alpha + 1)\)
                                        & \(2 \ell \cdot d \cdot (d \cdot \alpha + 1)\) \\
        
        \textbf{Attention}              & \(8\ell \cdot d^2 + 2 \ell^2 \cdot h \cdot (2d + 1)\)
                                        & \(\ell \cdot d \cdot (4d + 1) + \ell^2 \cdot h \cdot (2d + 1)\)
                                        & \(4\ell \cdot d^2 + \ell^2 \cdot h (2d + 3)\)\\
        
        \textbf{Efficient Attention}    & \(4\ell \cdot d^2 + 4\ell^2 \cdot d + 2\ell^2\)    
                                        & \(2\ell \cdot d^2 + 2\ell^2 \cdot d + \ell^2 + \ell \cdot d \)
                                        & \(2\ell \cdot d^2 + 2\ell^2 \cdot d + 2\ell^2\)\\        
                                        
        \textbf{Eff Diff Skip Attention}& \(\ell \cdot d (4d + 4 \ell +2d\cdot \alpha + k + 4) +2\ell^2\) 
                                        & \(\ell \cdot d (2d + 2\ell + d \cdot \alpha + k + 3) + \ell^2\)
                                        & \(\ell \cdot d (2d + 2\ell + d \cdot \alpha + k + 7) + \ell^2\) \\
                                        
        \textbf{MAMBA main}             & \(\ell \cdot i \cdot (d \cdot 6 + 9) + \ell \cdot d + c + SSM\)     
                                        & \(\ell \cdot i \cdot (d \cdot 3 + 2 + c) + \ell \cdot d + SSM\)     
                                        & \(\ell \cdot i \cdot (d \cdot 3 + 5 + c) + SSM\) \\

        \textbf{SSM}                    & \(\ell \cdot i \cdot (d_s \cdot 18 + \rho \cdot 4 + 8) + i \)
                                        & \(\ell \cdot i \cdot (d_s \cdot 4 + \rho \cdot 2 + 2)\)
                                        & \(\ell \cdot i \cdot (d_s \cdot 7 + \rho \cdot 2 + 2)\) \\
        \hline
        \end{tabular}
    }
    \label{table:complexity_appendix}
\end{table*}

\subsection{\sname{}}
This section focuses on analyzing the memory and computational costs of the Efficient Encoder block of \sname{}. Its first path consists in Efficient Attention, so we consider its weight count and activation size, and introduce the modifications due to the Convolutional Skip Connection and the weighted sum mechanisms.
For the weights, we include the contributions from the Convolutional and Fully Connected layers. However, the four vectors used for the weighted sum during training do not contribute to the final memory footprint, as they can be discarded and replaced by two weights computed at runtime. This results in:
\begin{equation}
    W_{EffEnc} = 2 d^2 + d^2 \cdot \alpha + k \cdot d \cdot \alpha   
\end{equation}

For the activations, we only need to consider the maximum between those resulting from the Efficient Attention and those from the Convolutional Skip component. The new component requires an activation size of at most \(d \cdot \ell (2 + \alpha)\), which arises from the processing of the fully connected layer and the attention result that must be retained in memory. This results in a total activation size of:
\begin{equation}
    A_{EffEnc} = \max ( 2 d \cdot \ell + \ell^2 ; \  d \cdot \ell (2 + \alpha ))
\end{equation}

For the computational complexity, we start with the operations required by the Efficient Attention and add those introduced by the Convolutional Skip Connection. The convolution step requires \(d \cdot l \cdot (k + 1)\) memory accesses, along with \(d \cdot l \cdot k\) sums and multiplications. 

Next, the SiLU activation requires~\(d \cdot l\) memory accesses, approximately \(d \cdot l\) summations, and \(4 d \cdot l \) multiplications. 

The fully connected layer introduces~\(2 d^2 \cdot \alpha \cdot \ell \) memory accesses, along with half as many summations and multiplications~(\(d^2 \cdot \alpha \cdot \ell \)).

Finally, the aggregation step requires~\(2 d \cdot \ell\) multiplications and memory accesses, with only half as many summations.

\subsection{Other blocks}
In this section, we synthetically review the two main other blocks required for a complete analysis of the Transformer/BERT architecture.

\paragraph{Feed Forward block}
This block, typically appended to the Attention or Efficient Attention layers after a normalization step, consists of two fully connected layers with size \(d^2 \cdot \alpha\). 
These layers sequentially increase/decrease the embedding dimension by a factor of\(\alpha\) with an activation function applied in between, followed by a skip connection. The total parameter count for this block is:
\begin{equation}
    W_{ff} = 2d^2 \cdot \alpha
\end{equation}
and activation size of: 
\begin{equation}
    A_{ff} = 2 \ell \cdot d + \ell \cdot d \cdot \alpha.
\end{equation}
Overall, the Feed Forward block requires~\(4 \ell \cdot d \cdot (d \cdot \alpha + 1)\) 
memory accesses and half as many summations and multiplications, with the majority of these operations required by the matrix multiplications in the fully connected layers.

\paragraph{Normalization Layer}
The normalization layer performs simple layer-wise normalization, requiring~\(2 \cdot d\) 
parameters to store the mean and variance values. During inference, it performs~\((\ell+1) \cdot d \cdot 2\) memory accesses and~\(\ell \cdot d\) 
summations and multiplications. The required activation size is~\(\ell \cdot d \cdot 2\), which is negligible compared to the activation sizes of the more computationally intensive layers.

\section{Complete results}
\label{appendix:completeresults}

\begin{table*}[htbp]
    \caption{Comparison of accuracy of pretrained \sname{} and \sname{}-Q models on the TinyNLP benchmark.}
    \begin{center}
        \begin{tabular}{| c | c c c c c c c | c |}
        \hline
        \textbf{Model} & \href{https://huggingface.co/datasets/stanfordnlp/imdb}{\textbf{IMDb}} & \href{https://huggingface.co/datasets/fancyzhx/ag_news}{\textbf{ag\_news}} & \href{https://www.kaggle.com/datasets/andrewmvd/cyberbullying-classification}{\textbf{cyberbull}} & \href{https://huggingface.co/datasets/IBM/limit}{\textbf{LiMiT}} & \href{https://huggingface.co/datasets/dair-ai/emotion}{\textbf{Emotion}} & \href{https://huggingface.co/datasets/xingkunliuxtracta/nlu_evaluation_data}{\textbf{nlu}} & \href{https://huggingface.co/datasets/benayas/snips}{\textbf{Snips}} & \textbf{Average} \\
        \hline \hline
        \textbf{\sname{}}            & 84,10 & 90,46 & 83,97 & 76,36 & 89,58 & 88,16 & 97,67 & 87,19 \\
        \textbf{\sname{}-Q}      & 84,01 & 90,63 & \textbf{86,60} & 74,10 & \textbf{89,90} & \textbf{94,05} & \textbf{97,93} & \textbf{88,17} \\
        \hline
        \end{tabular}
        \label{table:all_results_tinyNLP_acc_quant}
    \end{center} 
\end{table*}

\begin{table*}[htbp]
    \caption{Comparison of accuracy of pretrained \sname{} and \sname{}-Q models on the GLUE benchmark. We report SCC for STSB, MCC for CoLA, F1~score for QQP and MRPC, Accuracy for the remaining GLUE tasks.}
    \begin{center}
    \resizebox{\textwidth}{!}{
        \begin{tabular}{| c | c c c c c c c c c c | c |}
        \hline 
        \textbf{Model} & COLA & SST-2 & MRPC & QQP & MNLI-m & MNLI-mm & QNLI & RTE & WNLI & STSB & Score \\
        \hline \hline
        \textbf{\sname{}}            & \textbf{11,01} & 79,33 & \textbf{69,19} & 83,25 & \textbf{67,83} & \textbf{68,63} & \textbf{68,92} & 49,96 & \textbf{87,61} & 49,25 & \textbf{63,50} \\
        \textbf{\sname{}-Q}      & 10,66  & 80,96 & 67,99 & 82,45 & 67,10 & 68,05 & 68,06 & 47,29 & 87,32 & \textbf{49,28} & 63,11 \\
        \hline
        \end{tabular}

    \label{table:res_glue_comp_quant}
    }
    \end{center}    
\end{table*}

In this section, we provide the complete results for all our experiments, spanning all datasets and models in both pretrained and non-pretrained contexts. 

\subsection{Quantized models}

The complete results of the quantized and unquantized versions of \sname{} on the TinyNLP benchmark and the Glue Benchmark are reported in Figure \ref{table:all_results_tinyNLP_acc_quant} and Figure \ref{table:res_glue_comp_quant}, respectively.

\subsection{Scaled models}

The complete results of the scaled versions of \sname{} on the GLUE benchmark in both their floating point and quantized forms are reported in Figure \ref{table:GLUE_scaling_effect_fp} and Figure \ref{table:GLUE_scaling_effect_quant}, respectively.

\subsection{Evaluating the effect of the components}

\begin{table*}[tbp]
    \caption{Accuracy of the different pregressor models of \sname{} presented in the ablation on the TinyNLP benchmark.}
    \begin{center}
        \begin{tabular}{| c | c c c c c c c | c |}
        \hline
        \textbf{Model} & \href{https://huggingface.co/datasets/stanfordnlp/imdb}{\textbf{IMDb}} & \href{https://huggingface.co/datasets/fancyzhx/ag_news}{\textbf{ag\_news}} & \href{https://www.kaggle.com/datasets/andrewmvd/cyberbullying-classification}{\textbf{cyberbull}} & \href{https://huggingface.co/datasets/IBM/limit}{\textbf{LiMiT}} & \href{https://huggingface.co/datasets/dair-ai/emotion}{\textbf{Emotion}} & \href{https://huggingface.co/datasets/xingkunliuxtracta/nlu_evaluation_data}{\textbf{nlu}} & \href{https://huggingface.co/datasets/benayas/snips}{\textbf{Snips}} & \textbf{Average} \\
        \hline \hline
        BERT(2MB)           & 79,38 & 89,00 & 83,90 & 74,72 & 77,34 & 86,14 & 97,00 & 83,93 \\
        \hline \hline
        BERT~+~NE            & 81,86 & 89,40 & 81,38 & 74,72 & 45,72 & 70,10 & 96,40 & 77,08 \\
        BERT~+~EA            & 80,46 & 89,46 & 84,58 & 74,12 & 85,78 & 87,44 & 97,62 & 85,64 \\
        BERT~+~NE~+~EA         & 83,19 & 90,80 & 84,13 & 75,80 & 88,70 & 88,88 & 97,79 & 87,04 \\
        \hline \hline
        \textbf{\sname{}}            & 84,10 & 90,46 & 83,97 & 76,36 & 89,58 & 88,16 & 97,67 & 87,19 \\
        \hline
        \end{tabular}
        \label{table:all_results_tinyNLP_acc_abl}
    \end{center} 
\end{table*}

\begin{table*}[htbp]
    \caption{Accuracy of the different pregressor models of \sname{} presented in the ablation on the GLUE benchmark. We report SCC for STSB, MCC for CoLA, F1~score for QQP and MRPC, Accuracy for the remaining GLUE tasks.\mas{rewrite this}}
    \begin{center}
    \resizebox{\textwidth}{!}{
        \begin{tabular}{| c | c c c c c c c c c c | c |}
        \hline 
        \textbf{Model} & COLA & SST-2 & MRPC & QQP & MNLI-m & MNLI-mm & QNLI & RTE & WNLI & STSB & Score \\
        \hline \hline
        BERT(2MB)           & -0,86 & 71,28 & 64,66 & 73,04 & 60,56 & 61,58 & 60,82 & 48,24 & 66,20 & 15,48 & 52,10   \\
        \hline \hline
        BERT~+~NE            & 9,04  & 78,82 & 65,04 & 79,96 & 63,08 & 63,30 & 63,20 & 51,76 & 87,30 & 13,46 & 57,50 \\
        BERT~+~EA            & 10,06 & 79,44 & 66,48 & 78,88 & 60,82 & 60,82 & 63,50 & 50,76 & 22,24 & 17,92 & 51,09 \\
        BERT~+~NE~+~EA         & 18,70 & 79,60 & 65,36 & 82,66 & 67,06 & 67,44 & 67,44 & 53,66 & 86,74 & 23,34 & 61,20 \\
        \hline \hline
        \textbf{\sname{}}            & 11,01 & 79,33 & 69,19 & 83,25 & 67,83 & 68,63 & 68,92 & 49,96 & 87,61 & 49,25 & 63,50 \\
        \hline
        \end{tabular}

    \label{table:res_glue_abl}
    }
    \end{center}

\end{table*}

The complete results of the models in the ablation study on the TinyNLP benchmark and the Glue Benchmark are reported in Figure \ref{table:all_results_tinyNLP_acc_abl} and Figure \ref{table:res_glue_abl}, respectively.

\subsection{Evaluating the effect of pretraining}

\begin{table*}[tbp]
    \caption{Accuracy of non-pretrained models on the TinyNLP benchmark.}
    \begin{center}
        \begin{tabular}{| c | c c c c c c c | c |}
        \hline
        \textbf{Model} & \href{https://huggingface.co/datasets/stanfordnlp/imdb}{\textbf{IMDb}} & \href{https://huggingface.co/datasets/fancyzhx/ag_news}{\textbf{ag\_news}} & \href{https://www.kaggle.com/datasets/andrewmvd/cyberbullying-classification}{\textbf{cyberbull}} & \href{https://huggingface.co/datasets/IBM/limit}{\textbf{LiMiT}} & \href{https://huggingface.co/datasets/dair-ai/emotion}{\textbf{Emotion}} & \href{https://huggingface.co/datasets/xingkunliuxtracta/nlu_evaluation_data}{\textbf{nlu}} & \href{https://huggingface.co/datasets/benayas/snips}{\textbf{Snips}} & \textbf{Average} \\
        \hline \hline
        Embedder           & 82,60 & 91,10 & 82,78 & 71,60 & 89,40 & 89,50 & \textbf{97,93} & 86,41 \\
        Embedder~+~Conv      & 84,08 & \textbf{91,50} & 83,10 & 70,32 & 89,45 & 89,33 & 97,75 & 86,50 \\
        MAMBA(2MB)          & 78,18 & 91,08 & 83,74 & 71,66 & 77,40 & 87,60 & 97,34 & 83,86 \\
        BERT(2MB)           & 78,98 & 88,93 & 82,63 & 70,10 & 83,35 & 85,63 & 96,58 & 83,74 \\
        \hline \hline
        \textbf{\sname{}}            & 83,10 & 90,82 & 82,50 & 68,90 & 68,48 & 76,78 & 95,18 & 80,82 \\
        \hline
        \end{tabular}
        \label{table:all_results_tinyNLP_acc_unpretr}
    \end{center} 
\end{table*}

\begin{table*}[tbp]
    \caption{Accuracy of pretrained and finetuned models on the TinyNLP benchmark (Embedder and Embedder~+~Conv are directly trained on the downstream datasets).}
    \begin{center}
        \begin{tabular}{| c | c c c c c c c | c |}
        \hline
        \textbf{Model} & \href{https://huggingface.co/datasets/stanfordnlp/imdb}{\textbf{IMDb}} & \href{https://huggingface.co/datasets/fancyzhx/ag_news}{\textbf{ag\_news}} & \href{https://www.kaggle.com/datasets/andrewmvd/cyberbullying-classification}{\textbf{cyberbull}} & \href{https://huggingface.co/datasets/IBM/limit}{\textbf{LiMiT}} & \href{https://huggingface.co/datasets/dair-ai/emotion}{\textbf{Emotion}} & \href{https://huggingface.co/datasets/xingkunliuxtracta/nlu_evaluation_data}{\textbf{nlu}} & \href{https://huggingface.co/datasets/benayas/snips}{\textbf{Snips}} & \textbf{Average} \\
        \hline \hline
        MAMBA(2MB)          & 84,76 & 90,68 & 81,20 & 73,98 & 74,58 & 73,24 & 93,42 & 81,69 \\
        BERT(2MB)           & 79,38 & 89,00 & 83,90 & 74,72 & 77,34 & 86,14 & 97,00 & 83,93 \\
        \hline \hline
        \textbf{\sname{}}            & 84,10 & 90,46 & 83,97 & 76,36 & 89,58 & 88,16 & 97,67 & 87,19 \\
        \hline
        \end{tabular}
        \label{table:all_results_tinyNLP_acc}
    \end{center} 
\end{table*}

\begin{table*}[htbp]
    \caption{Evauation of non-pretrained models on the GLUE benchmark. We report SCC for STSB, MCC for CoLA, F1~score for QQP and MRPC, Accuracy for the remaining GLUE tasks.}
    \begin{center}
    \resizebox{\textwidth}{!}{
        \begin{tabular}{| c | c c c c c c c c c c | c |}
        \hline
        \textbf{Model} & COLA & SST-2 & MRPC & QQP & MNLI-m & MNLI-mm & QNLI & RTE & WNLI & STSB & Score \\
        \hline \hline
        Embedder           & 9,65 & 78,90 & 62,25 & 83,28 & 62,06 & 62,17 & 65,40 & 52,73 & 77,20 & 15,58 & 56,92 \\
        Embedder~+~Conv      & 9,25 & 79,10 & 60,50 & 82,98 & 61,98 & 60,93 & 62,08 & 52,00 & 79,16 & 16,10 & 56,41 \\
        MAMBA(2MB)         & 7,22 & 80,60 & 60,72 & 80,66 & 64,27 & 64,44 & 59,78 & 51,76 & 80,00 & 4,58 & 55,40 \\
        BERT(2MB)           & 3,93 & 75,20 & 63,88 & 81,03 & 64,95 & 63,65 & 62,75 & 49,38 & 70,68 & 6,74 & 54,22 \\
        \hline \hline
        \textbf{\sname{}}            & 5,32 & 78,50 & 62,54 & 82,58 & 63,82 & 65,78 & 63,68 & 51,26 & 87,30 & 9,76 & 57,05 \\
        \hline
        \end{tabular}

    \label{table:res_glue_comp_unpretr}
    }
    \end{center}

\end{table*}

\begin{table*}[htbp]
    \caption{Evaluation of pretrained and finetuned models on the GLUE benchmark (Embedder and Embedder~+~Conv are directly trained on the donwstream datasets). We report SCC for STSB, MCC for CoLA, F1~score for QQP and MRPC, Accuracy for the remaining GLUE tasks.}
    \begin{center}
    \resizebox{\textwidth}{!}{
        \begin{tabular}{| c | c c c c c c c c c c | c |}
        \hline 
        \textbf{Model} & COLA & SST-2 & MRPC & QQP & MNLI-m & MNLI-mm & QNLI & RTE & WNLI & STSB & Score \\
        \hline \hline
        MAMBA(2MB)          & 2,56  & \textbf{81,16} & 64,62 & 79,18 & 61,22 & 61,40 & 63,20 & 50,20 & 76,62 & 10,16 & 55,03 \\ 
        BERT(2MB)           & -0,86 & 71,28 & 64,66 & 73,04 & 60,56 & 61,58 & 60,82 & 48,24 & 66,20 & 15,48 & 52,10   \\
        \hline \hline  
        \textbf{\sname{}}            & \textbf{11,01} & 79,33 & \textbf{69,19} & 83,25 & \textbf{67,83} & \textbf{68,63} & \textbf{68,92} & 49,96 & \textbf{87,61} & 49,25 & \textbf{63,50} \\
        \hline
        \end{tabular}

    \label{table:res_glue_comp}
    }
    \end{center}

\end{table*}

\begin{table*}[htbp]
    \caption{Evaluation of the scaling performance of pretrained and finetuned models on the GLUE benchmark. We report SCC for STSB, MCC for CoLA, F1 score for QQP and MRPC, Accuracy for the remaining GLUE tasks.}
    \begin{center}
    \resizebox{\textwidth}{!}{
        \begin{tabular}{| c | c c c c c c c c c c | c |}
        \hline 
        \textbf{Model} & COLA & SST-2 & MRPC & QQP & MNLI-m & MNLI-mm & QNLI & RTE & WNLI & STSB & Score \\
        \hline \hline
        \textbf{EmbBERT-Nano} & 11,88 & 87,32 & 64,18 & 77,80 & 57,69 & 58,04 & 62,07 & 54,08 & 68,23 & 11,26 & 55,26 \\
        \textbf{EmbBERT-Tiny} & 7,95 & 75,07 & 65,11 & 81,24 & 60,35 & 61,85 & 67,30 & 51,84 & 72,39 & 27,71 & 57,10 \\
        \textbf{EmbBERT} & 11,01 & 79,33 & 69,19 & 83,25 & 67,83 & 68,63 & 68,92 & 49,96 & 87,61 & 49,25 & 63,50 \\
        \textbf{EmbBERT-Med} & 16,78 & 91,72 & 66,37 & 81,74 & 62,41 & 63,59 & 67,03 & 52,74 & 84,38 & 56,97 & 64,37 \\
        \textbf{EmbBERT-Big} & 15,94 & 80,79 & 66,28 & 83,30 & 65,61 & 67,02 & 72,18 & 51,44 & 84,51 & 68,25 & 65,53 \\
        \hline
        \end{tabular}
    \label{table:GLUE_scaling_effect_fp}
    }
    \end{center}    
\end{table*}

\begin{table*}[htbp]
    \caption{Evaluation of quantized models on the GLUE benchmark. We report SCC for STSB, MCC for CoLA, F1 score for QQP and MRPC, Accuracy for the remaining GLUE tasks.}
    \begin{center}
    \resizebox{\textwidth}{!}{
        \begin{tabular}{| c | c c c c c c c c c c | c |}
        \hline 
        \textbf{Model} & COLA & SST-2 & MRPC & QQP & MNLI-m & MNLI-mm & QNLI & RTE & WNLI & STSB & Score \\
        \hline \hline
        \textbf{EmbBERT-Nano-Q} & 7,54 & 77,06 & 65,96 & 78,87 & 59,55 & 60,51 & 62,84 & 52,71 & 64,79 & 12,69 & 54,25 \\
        \textbf{EmbBERT-Tiny-Q} & 7,21 & 75,34 & 67,23 & 81,92 & 61,47 & 62,12 & 68,15 & 48,74 & 74,65 & 31,38 & 57,82 \\
        \textbf{\sname{}-Q}   & 10,66  & 80,96 & 67,99 & 82,45 & 67,10 & 68,05 & 68,06 & 47,29 & 87,32 & 49,28 & 63,11 \\
        \textbf{EmbBERT-Med-Q}  & 7,06 & 81,62 & 65,37 & 82,96 & 64,69 & 65,02 & 68,70 & 47,92 & 85,92 & 52,19 & 62,15 \\
        \textbf{EmbBERT-Big-Q}  & 20,36 & 81,82 & 65,14 & 84,11 & 67,54 & 68,31 & 73,77 & 48,56 & 84,51 & 69,02 & 66,31 \\
        \hline
        \end{tabular}
    \label{table:GLUE_scaling_effect_quant}
    }
    \end{center}    
\end{table*}


The complete results of the non-pretrained and pretrained models on the TinyNLP  Benchmarks are reported in Figure \ref{table:all_results_tinyNLP_acc_unpretr} and Figure \ref{table:all_results_tinyNLP_acc}, respectively; while the complete results of the non-pretrained and pretrained models on the GLUE Benchmarks are reported in Figure \ref{table:res_glue_comp_unpretr} and Figure \ref{table:res_glue_comp}, respectively.

\section{Parameters for models scaling}
\label{appendix:param_scaling}

\begin{table*}[tbp]
    \caption{\textbf{Model hyperparameters and memory sizes}. Columns represent: vocabulary size $v$, sentence length $\ell$, embedding dimension $d$, reduced embedding dimension $r_d$, forward expansion $\alpha$, number of attention heads $h$, SSM/convolution state size $d_s$, and number of layers. The last columns show the parameter, activation, total memory sizes, and the quantized dimension. \mas{Riscrivi questo.}}
    \begin{center}
        \begin{tabular}{|c|c c c c c c c c|}
            \hline
            \textbf{Model} & $v$ & $\ell$ & $d$ & $r_d$ & $\alpha$ & $h$ & $d_s$ & $N$ \\
            \hline \hline
            \textbf{EmbBERT-Nano}  & 2048  & 256 & 64  & 16 & 1 & 1 & 8   & 2   \\
            \hline
            \textbf{EmbBERT-Micro} & 2048  & 256 & 128 & 16 & 1 & 1 & 16  & 2  \\
            \hline
            \textbf{EmbBERT}       & 8192  & 256 & 128 & 16 & 1 & 1 & 32 & 4  \\
            \hline
            \textbf{EmbBERT-Med}   & 8192  & 256 & 256 & 32 & 2 & 1 & 64  & 6  \\
            \hline
            \textbf{EmbBERT-Big}   & 16384 & 1024 & 512 & 32 & 2 & 1 & 64  & 6   \\
            \hline
        \end{tabular}
    \end{center}
    \label{table:models_params_scaling_params}
\end{table*}

The scaling of the EmbBERT architecture was carried out through an iterative process guided by memory profiling and empirical evaluation rather than fixed proportional rules. Starting from the base \textit{EmbBERT} configuration ($\approx$2~MB total), each variant was derived by progressively modifying the parameters that most strongly affected memory while monitoring their impact on accuracy and computational cost. Analytical inspection of parameter contributions showed that the embedding dimension and the number of layers dominated the memory footprint, while activations were mainly influenced by the vocabulary size and sentence length. Consequently, the first step in downscaling was to reduce the vocabulary size and, when necessary, the embedding dimension, as these provided the largest memory savings with relatively contained accuracy loss. Further compression was achieved by slightly reducing the number of layers and keeping the reduced embedding dimension~$r_d$ small but constant to avoid the diminishing returns that limit the model’s internal representational capacity.

Table~\ref{table:models_params_scaling_params} summarizes the resulting configurations and highlights how each hyperparameter evolves across model sizes. The smaller variants, \textit{EmbBERT-Nano} and \textit{EmbBERT-Micro}, adopt minimal vocabularies $v$ (2048 tokens) and low embedding dimensions $d$ (64–128) with only two layers, achieving total sizes below 1~MB. The reference \textit{EmbBERT} balances expressiveness and compactness, maintaining a larger vocabulary $v$ (8192 tokens) and moderate depth $N$ (four layers) within a 2~MB budget. The medium and large models progressively increase both width and depth—doubling $d$ and $r_d$, expanding the forward factor~$\alpha$, and lengthening the sequence window—to enhance capacity at the cost of much higher activations sizes. Each configuration represents a calibrated trade-off derived from iterative profiling and small-scale validation, allowing the EmbBERT family to scale smoothly from ultra-compact variants suitable for embedded systems to larger, high-capacity versions for resource-rich environments.

\end{document}